\documentclass[twoside,11pt]{article}
\pdfoutput=1
\usepackage{jair, theapa, rawfonts}

\jairheading{1}{2015}{xx-xx}{x/xx}{x/xx}
\ShortHeadings{Scalable Bayesian Rule Lists}
{Yang, Rudin, \& Seltzer}
\firstpageno{1}

\usepackage{bm}
\usepackage{amsfonts}
\usepackage{amsmath} 
\usepackage{amssymb} 
\usepackage{graphicx} 
\usepackage{subfigure} 
\usepackage{verbatim}
\usepackage{algorithm}
\usepackage{algorithmic}

\newcommand*{\qed}{\hfill\ensuremath{\square}}
\usepackage{wrapfig}
\usepackage{framed}
\usepackage{tabularx}
\usepackage{multirow}
\usepackage{hhline}
\usepackage{booktabs}
\usepackage{bbm}
\usepackage{enumerate}
\usepackage{array}
\newcolumntype{L}{>{\centering}m{1.5cm}}
\newcolumntype{T}{>{\centering}m{1.5cm}}
\usepackage[justification=centering]{caption}
\usepackage[implicit=false]{hyperref}
\usepackage{color}

\newtheorem{definition}{Definition}
\newtheorem{theorem}{Theorem}
\newtheorem{proof}{Proof}
\newtheorem{lemma}{Lemma}
\newcolumntype{L}{>{\raggedright\arraybackslash}m{6.8cm}}

\def\posterior{\textrm{\rm Posterior}}
\def\likelihood{\textrm{\rm Likelihood}}
\def\prior{\textrm{\rm Prior}}
\def\fixed{\textrm{\rm better}}

\def\Captr{\textrm{Captr}}
\def\consolidated{\textrm{\rm consolidated}}
\def\hyp{\textrm{hypothetical}}
\begin{document}

\title{Scalable Bayesian Rule Lists}

\author{\name Hongyu Yang \email hongyuy@mit.edu \\
        \addr Department of Electrical Engineering and Computer Science \\
        Massachusetts Institute of Technology, USA 
        \AND
       \name Cynthia Rudin \email cynthia@cs.duke.edu \\
       \addr Department of Computer Science and Department of Electrical and Computer Engineering \\
       Duke University, USA
        \AND
       \name Margo Seltzer \email margo@eecs.harvard.edu\\
       \addr 
       School of Engineering and Applied Sciences\\
       Harvard University, USA
       }

\maketitle

\begin{abstract} 
We present an algorithm for building probabilistic rule lists that is two orders of magnitude faster than previous work. Rule list algorithms are competitors for decision tree algorithms. They are associative classifiers, in that they are built from pre-mined association rules. They have a logical structure that is a sequence of IF-THEN rules, identical to a decision list or one-sided decision tree. Instead of using greedy splitting and pruning like decision tree algorithms, we fully optimize over rule lists, striking a practical balance between accuracy, interpretability, and computational speed. The algorithm presented here uses a mixture of theoretical bounds (tight enough to have practical implications as a screening or bounding procedure), computational reuse, and highly tuned language libraries to achieve computational efficiency. Currently, for many practical problems, this method achieves better accuracy and sparsity than decision trees; further, in many cases, the computational time is practical and often less than that of decision trees. The result is a probabilistic classifier  (which estimates $P(y=1|\textbf{x})$ for each $\mathbf{x}$) that optimizes the posterior of a Bayesian hierarchical model over rule lists.
\end{abstract}

\section{Introduction}
\label{SecIntro}
Our goal is to build a competitor for decision tree algorithms in terms of accuracy, interpretability, and computational speed. Decision trees are widely used, particularly in industry, because of their interpretability. Their logical IF-THEN structure allows predictions to be explained to users. However, decision tree algorithms have the serious flaw that they are constructed using greedy splitting from the top down. They also use greedy pruning of nodes. They do not globally optimize any function, instead they are composed entirely of local optimization heuristics. If the algorithm makes a mistake in the splitting near the top of the tree, it is difficult to undo it, and consequently the trees become long and uninterpretable, unless they are heavily pruned, in which case accuracy suffers. In general, decision tree algorithms are computationally tractable, not particularly accurate, and less sparse and interpretable than they could be. This leaves users with no good alternative if they desire an accurate yet sparse logical classifier.

The method provided here provides probabilistic predictions, which means it aims to accurately predict $P(y=1|\textbf{x})$ for each $\mathbf{x}$. Several important ingredients provide the underpinning for our method including: 

\begin{enumerate}
\item[(i)] A \textbf{principled objective}, which is the posterior distribution for the Bayesian Rule List (BRL) model of \shortciteS{LethamRuMcMa15}. We optimize this objective over rule lists. Our algorithm is called Scalable Bayesian Rule Lists (SBRL).
\item[(ii)] A useful \textbf{statistical approximation} that narrows the search space. We assume that each leaf of the rule list contains (``captures'') a number of observations that is bounded below. Because of this approximation, the set of conditions defining each leaf is a frequent pattern. This means the antecedents within the rule list are all frequent patterns. All of the possible frequent patterns can be pre-mined from the dataset using one of the standard frequent pattern mining methods. This leaves us with a much smaller optimization problem: we optimize over the set of possible pre-mined rules and their order to create the rule list.
\item[(iii)] \textbf{High performance language libraries} to achieve computational efficiency. Optimization over rule lists can be solved by repeated low level computations that have the capacity to be sped up. At every iteration, we make a change to the rule list and need to evaluate the new rule list on the data. The high performance calculations speed up this evaluation.
\item[(iv)]  \textbf{Computational reuse}. When we evaluate a rule list on the data that has been modified from a previous rule list, we need only to change the evaluation of points below the change in the rule list. Thus we can reuse the computation above the change. 
\item[(v)] \textbf{Analytical bounds} on BRL's posterior that are tight enough to be used in practice for screening association rules and providing bounds on the optimal solution. These are provided in two theorems in this paper.
\end{enumerate}


Through a series of controlled experiments, we show that SBRL is over two orders of magnitude faster than the previous best code for this problem. 

Let us provide some sample results.
Figure \ref{FigMushroomExample} presents an example of a rule list that we learned for the UCI mushroom dataset \cite<see>{Bache+Lichman:2013}.
This rule list is a predictive model for whether a mushroom is edible. It was created in about 9 seconds on a laptop and achieves perfect out-of-sample accuracy. Figure \ref{FigAdultExample} presents a rule list for the UCI adult dataset \cite<see>{Bache+Lichman:2013}. We ran our SBRL algorithm for approximately 18 seconds on a laptop to produce this. The algorithm achieves a higher out-of-sample AUC (area under the ROC Curve) than that achieved if CART or C4.5 were heavily tuned on the test set itself.
\begin{figure} [htb]
\centering
\resizebox{\columnwidth}{!}{%
\begin{tabular}{rLl}
\textrm{if }	&	\textrm{bruises=no, odor=not-in-(none,foul)}	&	\textrm{ then probability that the mushroom is edible } =	0.00112	\\
\textrm{else if }	&	\textrm{odor=foul, gill-attachment=free},	&	\textrm{ then probability that the mushroom is edible } =	0.0007 	\\
\textrm{else if }	&	\textrm{gill-size=broad, ring-number=one},	&	\textrm{ then probability that the mushroom is edible } =	0.999 	\\
\textrm{else if }	&	\textrm{stalk-root=unknown, stalk-surface-above-ring=smooth},	&	\textrm{ then probability that the mushroom is edible } =	0.996 	\\
\textrm{else if }	&	\textrm{stalk-root=unknown, ring-number=one},	&	\textrm{ then probability that the mushroom is edible } =	0.0385 	\\
\textrm{else if }	&	\textrm{bruises=foul, veil-color=white},	&	\textrm{ then probability that the mushroom is edible } =	0.995 	\\
\textrm{else if }	&	\textrm{stalk-shape=tapering, ring-number=one},	&	\textrm{ then probability that the mushroom is edible } =	0.986 	\\
\textrm{else if }	&	\textrm{habitat=paths},	&	\textrm{ then probability that the mushroom is edible } =	0.958 	\\
\textrm{else }	&	\textrm{(default rule)}	&	\textrm{ then probability that the mushroom is edible } =	0.001  	\\
\end{tabular}%
}
\caption{Rule list for the mushroom dataset from the UCI repository \cite<data available from>{Bache+Lichman:2013}. \label{FigMushroomExample}}
\end{figure}

\begin{figure}
\centering
\begin{tabular}{rlc}
\textrm{if }	&	\textrm{capital-gain}$>$\$7298.00	&	\textrm{ then probability to make over 50K } =	0.986	\\
\textrm{else if }	&	\textrm{Young,Never-married},	&	\textrm{ then probability to make over 50K } =	0.003	\\
\textrm{else if }	&	\textrm{Grad-school,Married},	&	\textrm{ then probability to make over 50K } =	0.748	\\
\textrm{else if }	&	\textrm{Young,capital-loss=0},	&	\textrm{ then probability to make over 50K } =	0.072	\\
\textrm{else if }	&	\textrm{Own-child,Never-married},	&	\textrm{ then probability to make over 50K } =	0.015	\\
\textrm{else if }	&	\textrm{Bachelors,Married},	&	\textrm{ then probability to make over 50K } =	0.655	\\
\textrm{else if }	&	\textrm{Bachelors,Over-time},	&	\textrm{ then probability to make over 50K } =	0.255	\\
\textrm{else if }	&	\textrm{Exec-managerial,Married},	&	\textrm{ then probability to make over 50K } =	0.531	\\
\textrm{else if }	&	\textrm{Married,HS-grad},	&	\textrm{ then probability to make over 50K } =	0.300	\\
\textrm{else if }	&	\textrm{Grad-school},	&	\textrm{ then probability to make over 50K } =	0.266	\\
\textrm{else if }	&	\textrm{Some-college,Married},	&	\textrm{ then probability to make over 50K } =	0.410	\\
\textrm{else if }	&	\textrm{Prof-specialty,Married},	&	\textrm{ then probability to make over 50K } =	0.713	\\
\textrm{else if }	&	\textrm{Assoc-degree,Married},	&	\textrm{ then probability to make over 50K } =	0.420	\\
\textrm{else if }	&	\textrm{Part-time},	&	\textrm{ then probability to make over 50K } =	0.013	\\
\textrm{else if }	&	\textrm{Husband},	&	\textrm{ then probability to make over 50K } =	0.126	\\
\textrm{else if }	&	\textrm{Prof-specialty},	&	\textrm{ then probability to make over 50K } =	0.148	\\
\textrm{else if }	&	\textrm{Exec-managerial,Male},	&	\textrm{ then probability to make over 50K } =	0.193	\\
\textrm{else if }	&	\textrm{Full-time,Private},	&	\textrm{ then probability to make over 50K } =	0.026	\\
\textrm{else }	&	\textrm{(default rule)}	&	\textrm{ then probability to make over 50K } =	0.066.	
\end{tabular}
\caption{Rule list for the adult dataset from the UCI repository \cite<see>{Bache+Lichman:2013}. 
\label{FigAdultExample}}
\end{figure}


\section{Review of Bayesian Rule Lists of \shortciteA{LethamRuMcMa15}}
\label{SecBRLRecap}
Scalable Rule Lists uses the posterior distribution of the Bayesian Rule Lists algorithm. 
Our training set is $\{(x_i,y_i)\}_{i=1}^n$ where the $x_i\in\mathcal{X}$ encode features, and $y_i$ are labels, which in our case are binary, either $0$ or $1$.  A Bayesian decision list has the following form:
\begin{flushleft}
\noindent \textbf{if}\;\;\;\;\;\;\; $x$ obeys $a_1$ \textbf{then} $y \sim \textrm{Binomial}(\mathbf{\theta}_1)$, $\mathbf{\theta}_1 \sim \textrm{Beta}({\bm\alpha}+\mathbf{N}_1)$\\
\textbf{else if} $x$ obeys $a_2$ \textbf{then} $y \sim \textrm{Binomial}(\mathbf{\theta}_2)$, $\mathbf{\theta}_2 \sim \textrm{Beta}(\bm\alpha+{\mathbf{N}}_2)$\\
$\vdots$\\
\textbf{else if} $x$ obeys $a_m$ \textbf{then} $y \sim \textrm{Binomial}(\mathbf{\theta}_m)$, $\mathbf{\theta}_m \sim \textrm{Beta}({\bm\alpha}+\mathbf{N}_m)$\\
\textbf{else} $y \sim \textrm{Binomial}(\mathbf{\theta}_{0})$, $\mathbf{\theta}_{0} \sim \textrm{Beta}(\mathbf{\bm\alpha}+\mathbf{N}_0)$.
\end{flushleft}
Here, the antecedents $\{a_j\}_{j=1}^m$ are conditions on the $x$'s that are either true or false, for instance, if $x$ is a patient, $a_j$ is true when $x$'s age is above 60 years old and $x$ has diabetes, otherwise false. The vector $\bm\alpha=[\alpha_1,\alpha_0]$ has a prior parameter for each of the two labels. Values $\alpha_1$ and $\alpha_0$ are prior parameters, in the sense that each rule's prediction $y\sim  \textrm{Binomial}(\mathbf{\theta}_j)$, and $\mathbf{\theta}_j | {\bm\alpha} \sim \textrm{Beta}(\bm\alpha)$. The notation $\mathbf{N}_j$ is the vector of counts, where $N_{j,l}$ is the number of observations $x_i$ that satisfy condition $a_j$ but none of the previous conditions $a_1,...,a_{j-1}$, and that have label $y_i=l$, where $l$ is either 1 or 0.  $\mathbf{N}_j$ is added to the prior parameters $\bm\alpha$ from the usual derivation of the posterior for the Beta-binomial. The default rule is at the bottom, which makes predictions for observations that are not satisfied by any of the conditions. When an observation satisfies condition $a_j$ but not $a_1,...,a_{j-1}$ we say that the observation is \textit{captured} by rule $j$. Formally: 
\begin{definition}
Rule $j$ \textbf{captures} observation $i$, denoted $\Captr(i)=j$, when $$j=\textrm{argmin } j' \textrm{ such that } a_{j'}(x_i)=\textrm{True}.$$ 
\end{definition}

Bayesian Rule Lists is an associative classification method, in the sense that the antecedents are first mined from the database, and then the set of rules and their order are learned. The rule mining step is fast, and there are fast parallel implementations available. Any frequent pattern mining method will suffice, since the method needs only to produce all conditions with sufficiently high support in the database. The support of antecedent $a_j$ is denoted supp($a_j$), which is the number of observations that obey condition $a_j$. A condition is a conjunction of expressions ``feature$\in$values," e.g., age$\in$[40,50] and color=white.
The hard part is learning the rule list, which is what this paper focuses on.

The likelihood for the model discussed above is:
\[\likelihood = p(\mathbf{y}|\mathbf{x},d,\mathbf{\alpha})
\propto \prod_{j=0}^m \frac{  \Gamma (N_{j,0}+\alpha_0)\Gamma (N_{j,1}+\alpha_1)}{ \Gamma(N_{j,0} + N_{j,1} + \alpha_0+ \alpha_1)},\]
where $d$ denotes the rules in the list and their order, $d=(m,\{a_j,\theta_j\}_{j=0}^m)$.
Intuitively, one can see that having more of one class and less of the other class will make the likelihood larger. To see this, note that if $N_{j,0}$ is large and $N_{j,1}$ is small (or vice versa) the likelihood for rule $j$ is large.

Let us discuss the prior. There are three terms in the prior, one governing the number of rules $m$ in the list, one governing the size $c_j$ of each rule $j$ (the number of conditions in the rule), and one governing the choice of antecedent condition $a_j$ of rule $j$ given its size. Notation $a_{<j}$ includes the antecedents before $j$ in the rule list if there are any, \textit{e.g.}, $a_{<4} = \{a_1, a_2,a_3\}$. Also $c_{j}$ is the cardinality of antecedent $a_{j}$, also written $|a_{j}|$, as the number of conjunctive clauses in rule $a_j$. E.g., the rule `$x_{1}$=green' and `$x_{2}$\textless50' has cardinality 2. $c_{<j}$ includes the cardinalities of the antecedents before $j$ in the rule list. Notation $\mathcal{A}$ is the set of pre-mined antecedents. The prior is:
\begin{equation}\label{eq:priorprob}
\textrm{prior}(d|\mathcal{A},\lambda,\eta)=p(d|\mathcal{A},\lambda,\eta) = p(m|\mathcal{A},\lambda) \prod_{j=1}^m p(c_j|c_{<j},\mathcal{A},\eta) p(a_j|a_{<j},c_j,\mathcal{A}).
\end{equation}
The first term is the prior for the number of rules in the list.
Here, the number of rules $m$ is Poisson, truncated at the total number of pre-selected antecedents:
\begin{equation*}
p(m|\mathcal{A},\lambda) = \frac{(\lambda^m/m!)}{\sum_{j=0}^{|\mathcal{A}|} (\lambda^j/j!)}, \quad m=0,\ldots,|\mathcal{A}|,
\end{equation*}
where $\lambda$ is a hyper-parameter.
The second term in the prior governs the number of conditions in each rule.
The size of rule $j$ is $c_j$ which is Poisson, truncated to remove values for which no rules are available with that cardinality:
\begin{equation*}
p(c_j|c_{<j},\mathcal{A},\eta) = \frac{(\eta^{c_j}/c_j!)}{\sum_{k \in R_{j-1}(c_{<j},\mathcal{A})}  (\eta^k/k!)}, \quad c_j \in R_{j-1}(c_{<j},\mathcal{A}),
\end{equation*}
where $R_{j-1}$ is the set of cardinalities available after removing the first $j-1$ rules, and $\eta$ is a hyperparameter.
The third term in the prior governs the choice of antecedent, given that we have determined its size through the second term.
We simply have $a_j$ selected from a uniform distribution over antecedents in $\mathcal{A}$ of size $c_j$, excluding those in $a_{<j}$.
\begin{equation}\label{eq:uniformprior}
p(a_j|a_{<j},c_j,\mathcal{A}) \propto 1, \quad a_j \in Q_{c_j} = \{a \in \mathcal{A} \setminus \{a_1, a_2, ..., a_{j-1}\} : |a| = c_j\}.
\end{equation}
As usual, the posterior is proportional to the product of the likelihood and the prior.
\begin{equation*}
p(d|\mathbf{x},\mathbf{y},\mathcal{A},\mathbf{\alpha},\lambda,\eta) \propto p(\mathbf{y}|\mathbf{x},d,\mathbf{\alpha}) p(d|\mathcal{A},\lambda,\eta).
\end{equation*}
This is the full model, and the posterior $p(d|\mathbf{x},\mathbf{y},\mathcal{A},\mathbf{\alpha},\lambda,\eta)$ is what we aim to optimize to obtain the best rule lists. The hyperparameter $\lambda$ is chosen by the user to be the desired size of the rule list, and $\eta$ is chosen as the desired number of terms in each rule. The parameters $\alpha_0$ and $\alpha_1$ are usually chosen as 1 in order not to favor one class label over another.

Given the prior parameters $\lambda$, which governs the length of the list, $\eta$, which governs the desired number of conditions in the list, and $\alpha$, which provides a preference over labels (usually we set all the $\alpha$'s to 1), along with the set of pre-mined rules $\mathcal{A}$, the algorithm must select which rules from $\mathcal{A}$ to use, along with their order.

\section{Representation}
\label{SecSimAnneal}
We use an MCMC scheme: at each time $t$, we choose a neighboring rule list at random from the neighborhood by adding, removing, or swapping rules, building from the basic algorithm of \shortciteA{LethamRuMcMa15} as a starting point.
At each step, we need to evaluate the posterior function on each new rule list. Since this process is repeated many times during the algorithm, speeding up this particular subroutine can have a tremendous increase in computational speed. We improve the speed in three ways: we use high performance language libraries, computational reuse, and theoretical bounds, all of which are discussed in this section and the next.



\subsection{Expressing Computation as Bit Vectors}
The vast majority of the computational time spent constructing rule sets lies
in determining which rules \emph{capture} which observations in a particular rule ordering.
As a reminder, for a given ordering of rules in a set, we say that the first rule for
which an observation evaluates true captures that observation.
The na\"ive implementation of these operations calls for various set operations -- checking
whether a set contains an element, adding an element to a set, and removing an element
from a set.
However, set operations are typically slow, and hardware does little to help with
efficiency.

We convert all set operations to logical operations on bit vectors, for which hardware
support is readily available.
The bit vector representation is both memory- and computationally- efficient.
The vectors have length equal to the total number of data samples.
Before beginning the algorithm, for each rule, we compute the bit vector representing
the samples for which the rule generates a true value.
For a one million sample data set (or more precisely up to 1,048,576 observations) each rule carries with it 128 KB vector (since a byte consists of 8 bits), which fits comfortably in most L2 caches. 

For each rule list we consider, we maintain similarly sized vectors for each rule in
the set indicating which rule in the set captures which observation.
Within a rule list, each observation is captured by one and only one rule -- the first rule
for which the condition evaluates true.
Representing the rules and rule lists this way allows us to explore the rule list state space,
reusing significant computation.
 For example, consider a rule list containing $m$ rules.
Imagine that we wish to delete the $k^{th}$ rule from the set.
The na\"ive implementation recomputes the ``captures" vector for every rule in the set.
Our implementation updates only the captures vector for the rules 
at position $k$ or after position $k$ in the list, using logical operators acting upon
the rule list's ``captures" vector for rule $k$ and the rules that come after it.
This shortens the run time of the algorithm in practice by approximately 50\%.
This and other operations will be discussed in the next subsection. 

\subsection{An Algebra for Computation Reuse}

Our use of bit vectors transforms the large number of set operations (performed in a traditional implementation) into a set of boolean operations on bit vectors.
These customized operations are summarized below. In our notation, the rule list
contains $m$ rules; $a[j]$ and $a[k]$ are used to represent the $j^{th}$ and $k^{th}$ rules in the rule list. 

As a starting point, we define $a[k].captures$ as the captures vector for rule $a[k]$ that is computed during the course of computation. Define $a[k].init$
to be the original vector associated with each rule, indicating all observations for which the rule evaluates true.  Note that $a[k].captures \subset a[k].init$, because $a[k].captures$ can take value 1 only if $a[k].init$ is also 1. Below we show these bit vector operations for the possible MCMC steps. Here $\vee$ is the logical OR symbol and $\wedge$ is the logical AND symbol.
\begin{enumerate}
\item{
\begin{algorithmic}
	\mbox{Remove rule $a[k]$}
    \STATE $remaining \leftarrow a[k].captures$
	\FOR {$j = k+1$ to $m$}
		\STATE $tmp \leftarrow a[j].init \wedge remaining$	\COMMENT{Everything $a[k]$ previously captured that could be captured by rule $a[j]$}
		\STATE $a[j-1].captures \leftarrow a[j].captures \vee tmp$ \COMMENT{Include observations $a[k]$ previously captured that rule $a[j]$ now captures and move rule $a[j]$ up to the $j-1$ spot.}
		\STATE $remaining \leftarrow remaining \wedge \neg tmp$ \COMMENT{Remove the newly captured items from $remaining$} 
	\ENDFOR
    \STATE $m \leftarrow m-1$
\end{algorithmic}
}\vspace*{15pt}

\item{
\begin{algorithmic}
	\mbox{Insert rule into the ruleset at position $k$}
	\STATE shift rules $k$ to $m$ to positions $k+1$ to $m+1$ and insert rule at position $k$.
    \STATE $m \leftarrow m + 1$
	\STATE $captured \leftarrow \{\vec{0}_n\}$
	\STATE\COMMENT{The ``for" loop below counts what is captured up to position $k-1$}
	\FOR {$j$ = $1$ to $k - 1$ }
		\STATE $captured \leftarrow captured \vee j.captures$
	\ENDFOR
	\STATE\COMMENT{The ``for" loop below recomputes the captures vector for the rest of the rule list, which changed when we inserted the rule at position $k$.}
	\FOR {$j$ = $k$ to $m$}
		\STATE $a[j].captures \leftarrow a[j].init \wedge \neg{captured}$
		\STATE $captured \leftarrow a[j].captures \vee captured$
	\ENDFOR
\end{algorithmic}
}\vspace*{15pt}

\item{
\begin{algorithmic}
	\mbox{Generalized swap $a[j]$ and $a[k]$ ($j < k$) } \\
	\STATE $captured \leftarrow \{\vec{0}_n\}$ 
	\STATE \COMMENT{The ``for" loop below calculates observations captured by rules $a[j]$ through rule $a[k]$. Note that all of these observations will be captured again through new rule a[j] through new rule a[k].}
	\FOR{$t$ = $j$ to $k$}
		\STATE $captured \leftarrow captured \vee t.captures$
		 	\ENDFOR
	\STATE swap rules $j$ and $k$
	\STATE \COMMENT{The ``for" loop below recalculates observations captured by (swapped) rules $j$ through rule $k$.}
	\FOR{$t$ = $j$ to $k$ } 
		\STATE $a[t].captures \leftarrow captured \wedge a[t].init$
		\STATE $captured \leftarrow captured \wedge \neg{a[t].captures}$
	\ENDFOR
\end{algorithmic}
}\vspace*{15pt}
\end{enumerate}


\subsection{High Performance Bit Vector Manipulation and Ablation Study}

\begin{figure}
  \centering    \includegraphics[width=0.4
  \textwidth]{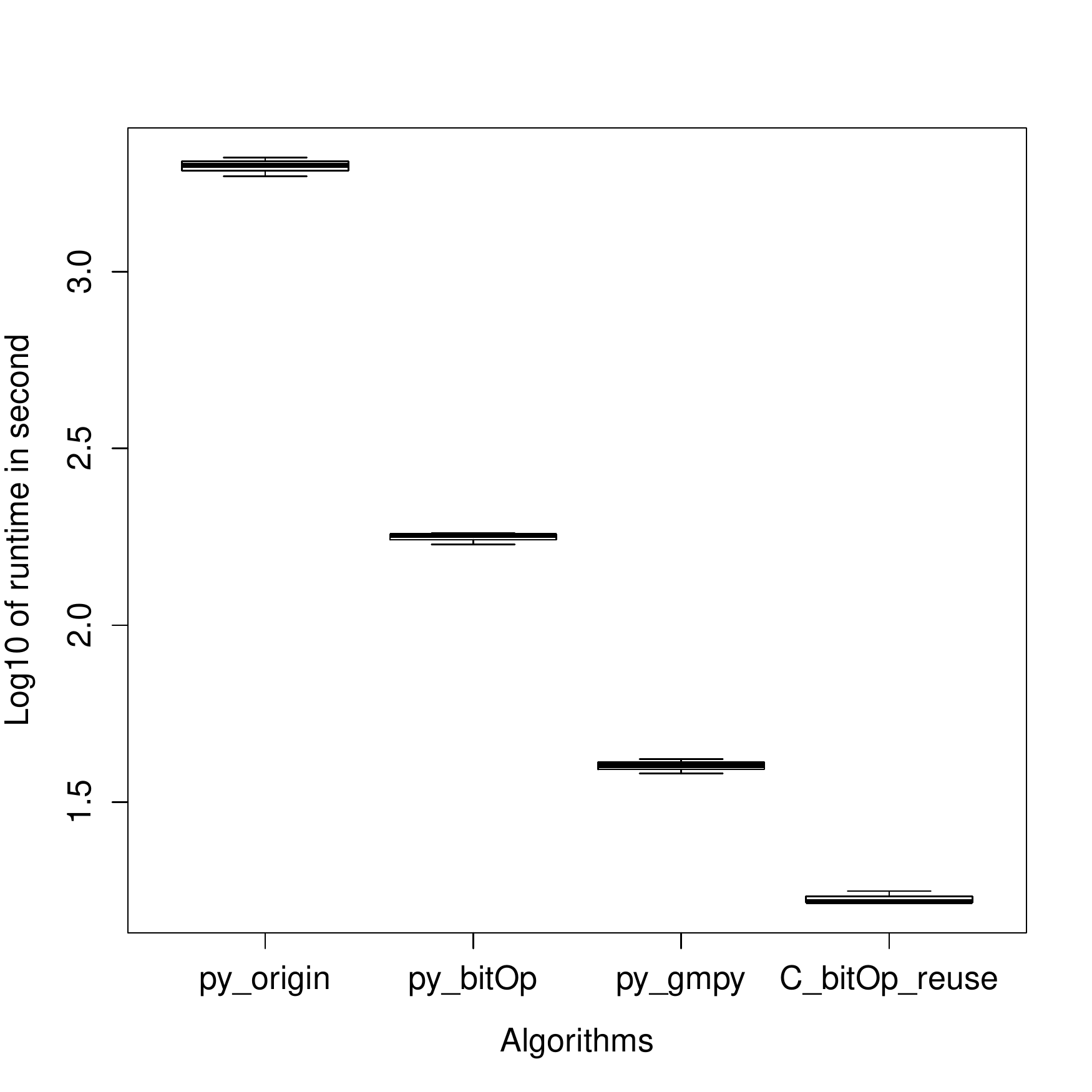}
  \caption{Boxplots of runtime comparison among different implementations. From the original python code, the final code is over two orders of magnitude faster.\label{boxplot_of_runtimecomparison}}
\end{figure} 

Having transformed expensive set operations into bit vector operations, we can now
leverage both hardware vector instructions and optimized software libraries.
We investigated four alternative implementations, each improving computational efficiency
from the previous one.
\begin{itemize}
\item First, we have the original python implementation here for comparison.
\item Next, we retained our python implementation but converted from set operations to bit operations.
\item Then, we used the python gmpy library to perform the bit operations.
\item Finally, we moved the implementation from Python to C, representing the bit vectors as multiprecision integers, using the
GMP library, which is faster on large data sets.
\end{itemize}

To evaluate how each of these steps improved the computation time of the algorithm, we conducted a controlled experiment where each version of the algorithm (corresponding to the four possibilities above) was given the same data (the UCI adult dataset, divided into ten folds), the same set of rules, and the same number of MCMC iterations (5,000 iterations for each of the 20 chains) to run. We created boxplots for the log10 of the run time over the different folds, which is shown in Figure \ref{boxplot_of_runtimecomparison}. The final code is over two orders of magnitude faster than the original optimized python code.


\section{Theoretical Bounds with Practical Implications\label{SecTheor}}

We prove two bounds. First we provide an upper bound on the number of rules in a maximum a posteriori rule list. This allows us to narrow our search space to rule lists below a certain size, if desired. Second we provide a constraint that states that certain prefixes can never lead to the maximum a posteriori rule list. This prevents our algorithm from searching in regions of the space that provably do not contain the maximum a posteriori rule list.

\subsection{Upper Bound on the Number of Rules in the List}


Given the number of features, the parameter $\lambda$ for the size of the list, and parameters $\alpha_0$ and $\alpha_1$, we can derive an upper bound for the size of a maximum a posteriori rule list. This formalizes how the prior on the number of rules is strong enough to overwhelm the likelihood.

We are considering binary rules and binary features, so the total number of possible rules of each size can be calculated directly. When creating the upper bound, within the proof, we hypothetically exhaust rules from each size category in turn, starting with the smallest sizes. We discuss this further below.

Let $|Q_{c}|$ be the number of antecedents that remain in the pile that have $c$ logical conditions. The sequence of $b$'s that we define next is a lower bound for the possible sequence of $|Q_{c}|$'s. In particular, $b$ represents the sequence of sizes of antecedents that would provide the smallest possible $|Q_{c}|$. Intuitively, the sequence of $b$'s arises when we deplete the antecedents of size 1, then deplete all of the antecedents of size 2, etc. The number of ways to do this is given exactly by the $b$ values, computed as follows.\\


\noindent \textbf{Definition}
Let $P$ be the number of features, and $\mathbf{b}=\left\{b_0, b_1, b_2,...b_{2^P-1}\right\}$ be a vector of length $2^P$ defined as follows: 
\begin{enumerate}
\item[]{
\begin{algorithmic}
    \textrm{index = 0}\\
	\FOR {$c = 0$ to $\left \lfloor{\frac{P}{2}}\right \rfloor $}
    	\STATE{
			\FOR {$j = \binom{P}{c}$ down to 1 (using step size=-1)}
            \STATE{
		$b_{\rm index}$ = j \\
            	index = index + 1 
                
            }
			\ENDFOR
        }
        \STATE{
        \IF{(c+c != $P$)}
        	\STATE{
			\FOR {$j = \binom{P}{P-c}$ down to 1 (using step size = -1)}
            \STATE{
		$b_{\rm index}$ = j \\
            	index = index + 1 
            }
			\ENDFOR
            }
        \ENDIF
        }
	\ENDFOR
\end{algorithmic}
}\vspace*{15pt}
\end{enumerate}

\begin{figure}
  \centering
 \includegraphics[width=0.55\textwidth]{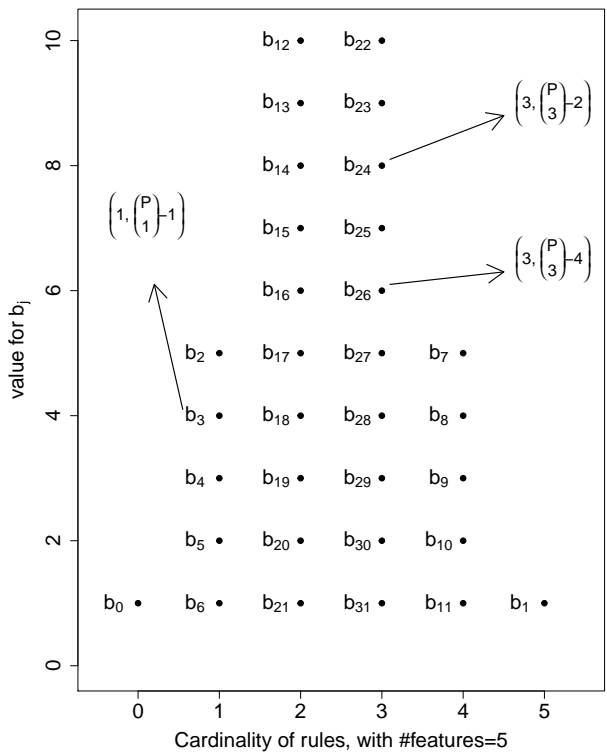}
  \caption{Example of $b_j$'s for a 5-feature dataset.\label{example_bjs}}
\end{figure}

Figure \ref{example_bjs} is an illustration for the $b_j$'s when the number of features is $P=5$.

We will use the $b$'s within the theorem below.
In our notation, rule list $d$ is defined by the antecedents and the probabilities on the right side of the rules,
$d=(m,\{a_l,\theta_l\}_{l=1}^m)$. 

\begin{theorem}\label{TheoremUpperBoundForm}
The size $m^*$ of any maximum a posteriori (MAP) rule list $d^*$ (with parameters $\lambda$, $\eta$, and $\alpha=(\alpha_0,\alpha_1)$) obeys $m^* \leq m_{max}$, where
\begin{align}
m_{\max}=\min \left\{ 2^P-1, \max \left\{ m' \in \mathbb{Z}_+ : \frac{\lambda^{m'}}{m'!} 
\geq 
\frac{\Gamma(N_-+\alpha_0)\Gamma(N_++\alpha_1)}{\Gamma(N+\alpha_0+\alpha_1)} 
\prod\limits_{j=1}^{m'} b_j\right\}\right\}.
\end{align}
In the common parameter choice $\alpha_0=1$ and $\alpha_1=1$, this reduces to:
\begin{align}
m_{\max}=\min \left\{ 2^P-1, \max \left\{ m' \in \mathbb{Z}_+ : \frac{\lambda^{m'}}{m'!} 
\geq 
\frac{\Gamma(N_-+1)\Gamma(N_++1)}{\Gamma(N+2)} 
\prod\limits_{j=1}^{m'} b_j\right\}\right\}.
\end{align}
\end{theorem}
The proof is in the appendix.

Figure \ref{figure_upperbound1} 
illustrates the use of this theorem. In particular, we plotted the upper bound for $m^*$ from the Theorem \ref{TheoremUpperBoundForm} when the number of features $P$ is 10 in Figure \ref{figure_upperbound1} (left), and we plotted the upper bound when the number of features is 15 in Figure \ref{figure_upperbound1} (right), with $\lambda=3$ and $\alpha_0=\alpha_1=1$. For instance, when there are 10 features (left plot) and approximately 100 positive and 100 negative observations, there will be at most about 36 rules. 

\begin{figure}
  \centering    \includegraphics[width=0.4\textwidth]{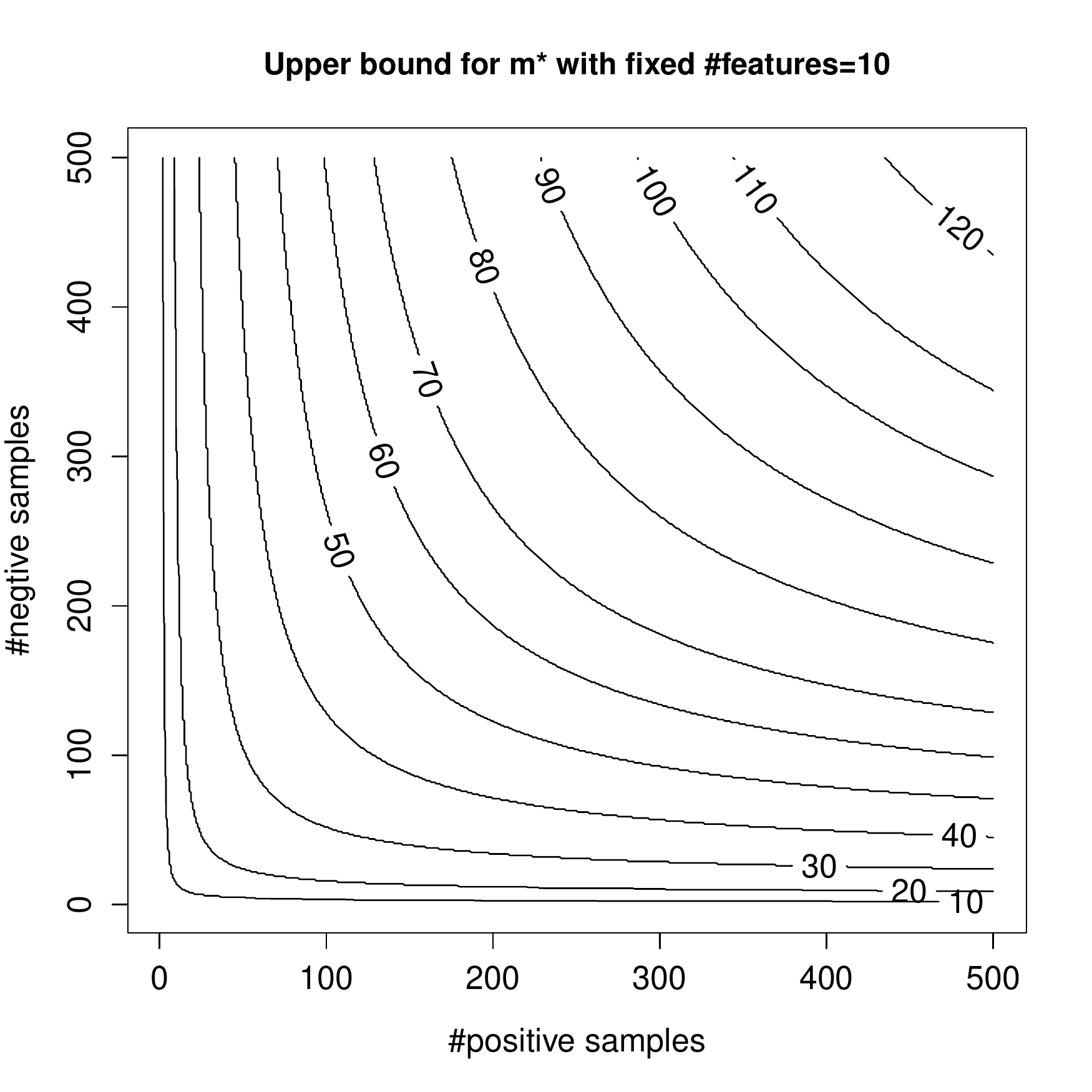}
 \includegraphics[width=0.4\textwidth]{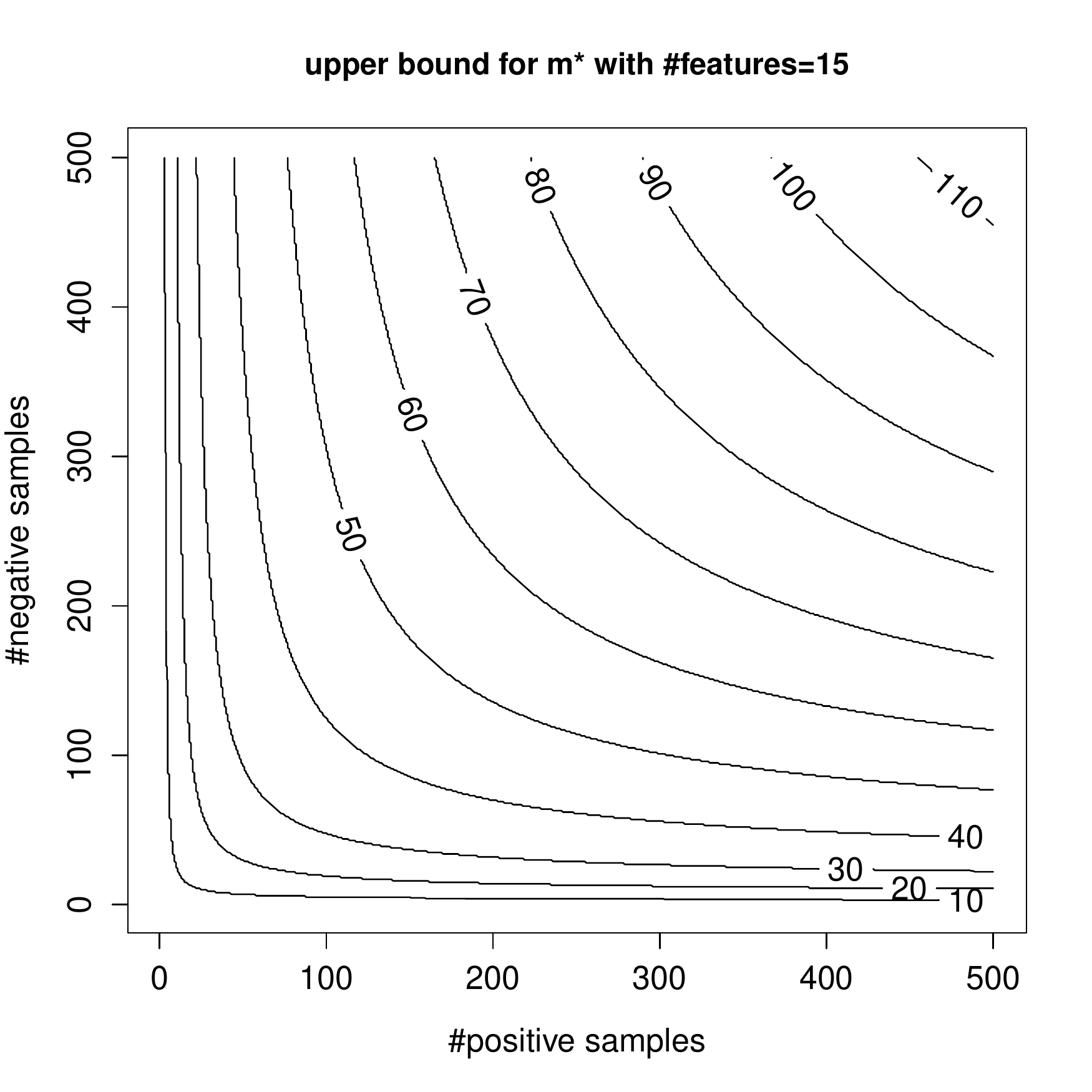}
  \caption{Upper bound from Theorem \ref{TheoremUpperBoundForm} for the length of the rule list when the number of features is 10 (left figure) and 15 (right figure). \label{figure_upperbound1}} 
\end{figure}


\subsection{Prefix Bound}
We next provide a bound that eliminates certain regions of the rule space from consideration. Consider a rule list beginning with rules $a_1,..,a_p$. If the best possible rule list starting with $a_1,..,a_p$ cannot beat the posterior of the best rule list we have found so far, then we know any rule list starting with $a_1,..,a_p$ is suboptimal. In that case, we should stop exploring rule lists starting with $a_1,..,a_p$. This is a type of branch and bound strategy, in that we have now eliminated (bounded) the entire set of lists starting with $a_1,..,a_p$. We formalize this intuition below.

\begin{figure}
  \centering    \includegraphics[width=0.4\textwidth]{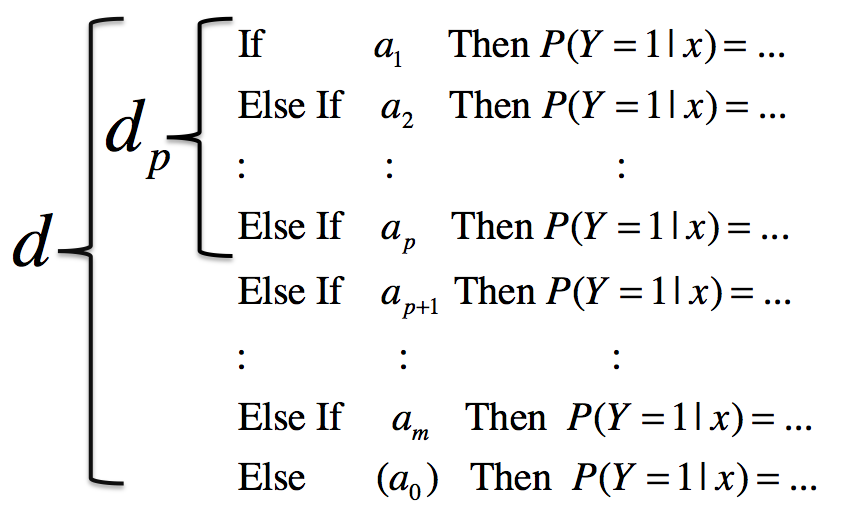}
  \caption{Notation for a rule list $d$ and its prefix $d_p$.\label{figuredp}}
\end{figure}

Denote the rule list at iteration t by $d^t = (a^t_1, a^t_2, ..., a^t_{m_t}, a_0)$. The current best posterior probability has value \(v^*_t\), that is $$v^*_t=\max_{t'\leq t} \posterior(d^{t'},\{(x_i,y_i)\}_{i=1}^n).$$ Let the current rule list be \(d = (a_1, a_2,...a_m, a_0)\). 
Let \(d_p\) denote a prefix of length $p$ of the rule list $d$, i.e., \(d_p=(a_1,a_2,...a_p)\), where \(a_1,a_2,...,a_p\) is the same as the first $p$ rules in $d$. Figure \ref{figuredp} illustrates this notation. We want to determine whether a rule list starting with $d_p$ could be better than the best we have seen so far.
Define $\Upsilon(d_p, \{(x_i,y_i)\}_{i=1}^n)$ as follows:
\begin{eqnarray*}
\lefteqn{ \Upsilon(d_p, \{(x_i,y_i)\}_{i=1}^n) }\\
&:=& 
\frac{\lambda^{\max{(p, \lambda)}}/{(\max{(p, \lambda)})}!}{\sum_{j=0}^{|\mathcal{A}|} (\lambda^j/j!)} \left(\prod_{j=1}^p p(c_j|c_{<j},\mathcal{A},\eta) \frac{1}{|Q_{c_j}|}\right) \times \\
&&\left(\prod_{j=0}^m \frac{  \Gamma (N_{j,0}+1)
\Gamma (N_{j,1}+1)
}{ \Gamma( N_{j,0}+ N_{j,1} + 2)}\right) 
\frac{ \Gamma (1+N_0-\sum_{j=1}^{p} N_{j,0}) }{ \Gamma(2+N_0-\sum_{j=1}^{p} N_{j,0})} \frac{ \Gamma (1+N_1-\sum_{j=1}^{p} N_{j,1}) }{ \Gamma(2+N_1-\sum_{j=1}^{p} N_{j,1})}.
\end{eqnarray*}
Here, $N_{j,0}$ is the number of points captured by rule $j$ with label 0, and $N_{j,1}$ is the number of points captured by rule $j$ with label 1,
\[
N_{j,0} = |\{i:\Captr(i)=j \textrm{ and } y_i=0\}|,\;\;N_{j,1}=|\{i:\Captr(i)=j \textrm{ and } y_i=1\}|.
\]

The result states that for a rule list with prefix $d_p$, if the upper bound on the posterior, $\Upsilon(d_p)$, is not as high as the posterior of the best rule list we have seen so far, then $d_p$ is a bad prefix, which cannot lead to a MAP solution. It tells us we no longer need to consider rule lists starting with $d_p$.
\begin{theorem}\label{Theorem_PrefixBound}
For rule list $d=\{d_p, a_{p+1}, ...,a_m, a_0\}$, if $$\Upsilon(d_p, \{(x_i,y_i)\}_{i=1}^n) < v^*_t,$$ then for $\alpha_0=1$ and $\alpha_1=1$, we have
\begin{align}
d
\not \in \textrm{\rm argmax}_{d'}\posterior(d', \{(x_i,y_i)\}_{i=1}^n).
\end{align}
\end{theorem}

Theorem \ref{Theorem_PrefixBound} is implemented in our code in the following way: for each random restart, the initial rule in the list is checked against the bound of Theorem \ref{Theorem_PrefixBound}. If the condition $\Upsilon(d_1)< v_t^*$ holds, we throw out this initial rule and choose a new one, because that rule provably cannot be the first rule in an optimal rule list. Theorem \ref{Theorem_PrefixBound} provides a substantial computational speedup in finding high quality or optimal solutions. In some cases, it provides a full order of magnitude speedup. Because it has been so useful in practice, we provide illustrative examples.

\subsection{Demonstrations of Theorem \ref{Theorem_PrefixBound}}\label{SubsectionPrefixBound}
\noindent\textbf{Demonstration 1:}
We use the tic tac toe dataset from the UCI repository \cite<see>{Bache+Lichman:2013}. Each observation is a tic tac toe board after the game has ended. If the $X$ player wins, the label of the observation is 1, otherwise it is 0.
Let us consider a rule list starting with the following two rules:
\begin{align*}
\textrm{If } \quad
&\begin{tabular}{ | l | c | r | }
  \hline
   &  &  \\ \hline
   &  & \\ \hline
   & o &  \\ \hline
\end{tabular} &
 \textrm{ then  ...} \\
 \textrm{ else if } \quad
&\begin{tabular}{ | l | c | r | }
  \hline
   &  &  \\ \hline
   &  & o\\ \hline
   &  &  \\ \hline
\end{tabular}&
 \textrm{ then  ...} \\
\textrm{else if ...}&&
\end{align*}
The first rule says that the board contains an ``O" in the bottom middle spot, and the rule says nothing about other spots. Intuitively this is a particularly bad rule, since it captures a lot of possible tic tac toe boards, and on its own, cannot distinguish between winning and losing boards for the ``X'' player. Similarly, the second rule also does not discriminate well. Thus, we expect any rule list starting with these two rules to perform poorly. We can show this using the theorem. On one of ten folds of the data, this rule list has a log posterior that is upper bounded at -272.51. From an earlier run of the algorithm, we know there is a rule list with a posterior of -105.012. (That rule list is provided in Table \ref{FigTicTac1}  and contains exactly one rule for each way the ``X'' player could have three X's in a row on the board.) Since the upper bound on the posterior for this rule list (-272.51) is less than -105.012, there does not exist an optimal rule list starting with these two rules.
\\

\noindent \textbf{Demonstration 2:}
In contrast with Demonstration 1, a rule list starting as follows cannot be excluded. 
\begin{align*}
\textrm{If } \quad
&\begin{tabular}{ | l | c | r | }
  \hline
   &  &  \\ \hline
  o & o & o \\ \hline
   &  &  \\ \hline
\end{tabular} 
& \textrm{ then  ...} \\
 \textrm{ else if } \quad
&\begin{tabular}{ | l | c | r | }
  \hline
  o &  &  \\ \hline
  o &  &  \\ \hline
  o &  &  \\ \hline
\end{tabular}
&
\textrm{ then  ...} \\
\textrm{else if ...}&&
\end{align*}
These first two rules says that the ``O'' player has three O's in a row, which means the ``X'' player could not have won. 
This prefix has a log posterior that is upper bounded at -35.90, which is higher than than -105.012. Thus we cannot exclude this prefix as being part of an optimal solution. As it turns out, there are high posterior solutions starting with this prefix. One such solution is shown in Table \ref{FigTicTac2} below.

\section{Experiments}
\label{SecExperiments}
We provide a comparison of algorithms along three dimensions: solution quality (AUC - area under the ROC curve), sparsity, and scalability. Sparsity will be measured as the number of leaves in a decision tree or as the number of rules in a rule list. 
Scalability will be measured in computation time. SBRL tends to achieve a useful balance between these three quantities.

Let us describe the experimental setup. As baselines, we chose popular classification algorithms to represent the sets of uninterpretable methods and the set of ``interpretable" methods. To represent the class of uninterpretable methods, we chose logistic regression, SVM RBF, random forests (RF), and boosted decision trees (ADA). None of these methods are designed to yield sparse classifiers. They are designed to yield scalable and accurate classifiers. To represent the class of ``interpretable" greedy splitting algorithms, we chose CART, C4.5, RIPPER, CBA, and CMAR. CART tends to yield sparse classifiers, whereas C4.5 tends to be much less interpretable. Other experiments \cite<see>{LethamRuMcMa15,WangRu15} have accuracy/interpretability comparisons to Bayesian Rule Lists and Falling Rule Lists, so our main effort here will be to add the scalability component. We used the RWeka R package's implementation of the RIPPERk algorithm, which is a generalized RIPPER algorithm with an optimization step repeated k times \cite<see>{Cohen95fasteffective}. For CBA, there are two implementations available: the original author's and one (LUCS-KDD) from Frans Coenen from University of Liverpool, which is recommended by the original author. For CMAR, the only available implementation referred to in multiple papers is the LUCS-KDD implementation. For convenience, we use the LUCS-KDD java implementations of TCV(10 cross validation) for both CBA and CMAR. The R implementation for the other algorithms are readily available in CRAN. We stored the experimental results for each dataset and each algorithm (R and java) and used this information to plot the figures and tables in this section. We benchmarked using publicly available datasets after data pre-processing (using quantiles to represent real-valued variables and merging discrete levels together to avoid too many levels in one column): 
\begin{itemize}
	\item the tic tac toe dataset \cite<see>{Bache+Lichman:2013}, where the goal is to determine whether the ``X'' player wins (this is easy for a human who would check for three X's in a row),
\item the adult dataset \cite<see>{Bache+Lichman:2013}, where we aim to predict whether an individual makes over \$50K in a year,
\item the mushroom dataset \cite<see>{Bache+Lichman:2013}, where the goal is to predict whether a mushroom is edible (as opposed to poisonous),
\item the nursery dataset \cite<see>{Bache+Lichman:2013}, where the goal is to predict whether a child's application to nursey school will be in either the ``very recommended'' or ``special priority'' categories,
\item the telco customer churn dataset (see \citeauthor{Telco}), where the goal is to predict whether a customer will leave the service provider,
\item the titanic dataset \cite<see>{Bache+Lichman:2013}, where the goal is to predict who survived the sinking of the Titanic.
\end{itemize}
Evaluations of prediction quality, sparsity, and timing were done using 10-fold cross validation.  

For creating the random starting rule list, the initial rule length was set to 1 rule (not including the default rule). The minimum and maximum rule size for the rule-mining algorithm were set at 1 and 2, respectively, except for the tic tac toe dataset, for which the maximum was set to 3. Because of the nature of the tic tac toe dataset, it is more interpretable to include rules of size 3 than to exclude them. For rule mining, we chose the minimum support of rules from (5\%, 10\%, 15\%, etc.) so that the total number of rules was approximately 300. 

The prior parameters were fixed at $\eta=1$, and $\alpha=(1,1)$.
For the $\lambda$ for each dataset, we first let $\lambda$ be 5, and ran SBRL once with the above parameters. Then we fixed $\lambda$ at the length of the returned rule list for that dataset, which is faster than nested cross-validation. For the purpose of providing a controlled experiment, the number of iterations was fixed at 5,000 for each chain of the 20 chains of SBRL, which we ran in series on a laptop. If the chains were computed in parallel rather than in series, it would speed up computation further. It is possible that the solution quality would increase if SBRL was run for a larger number of iterations. Every time SBRL started building a new rule list, we checked the initial rule in the list to see whether the upper bound on its posterior (by Theorem \ref{Theorem_PrefixBound}) was greater than the best rule list it had found so far. If not, the rule was replaced until the condition was satisfied.

\begin{figure}
\centering
\begin{minipage}{.5\textwidth}
  \centering
  \includegraphics[width=1.0\linewidth]{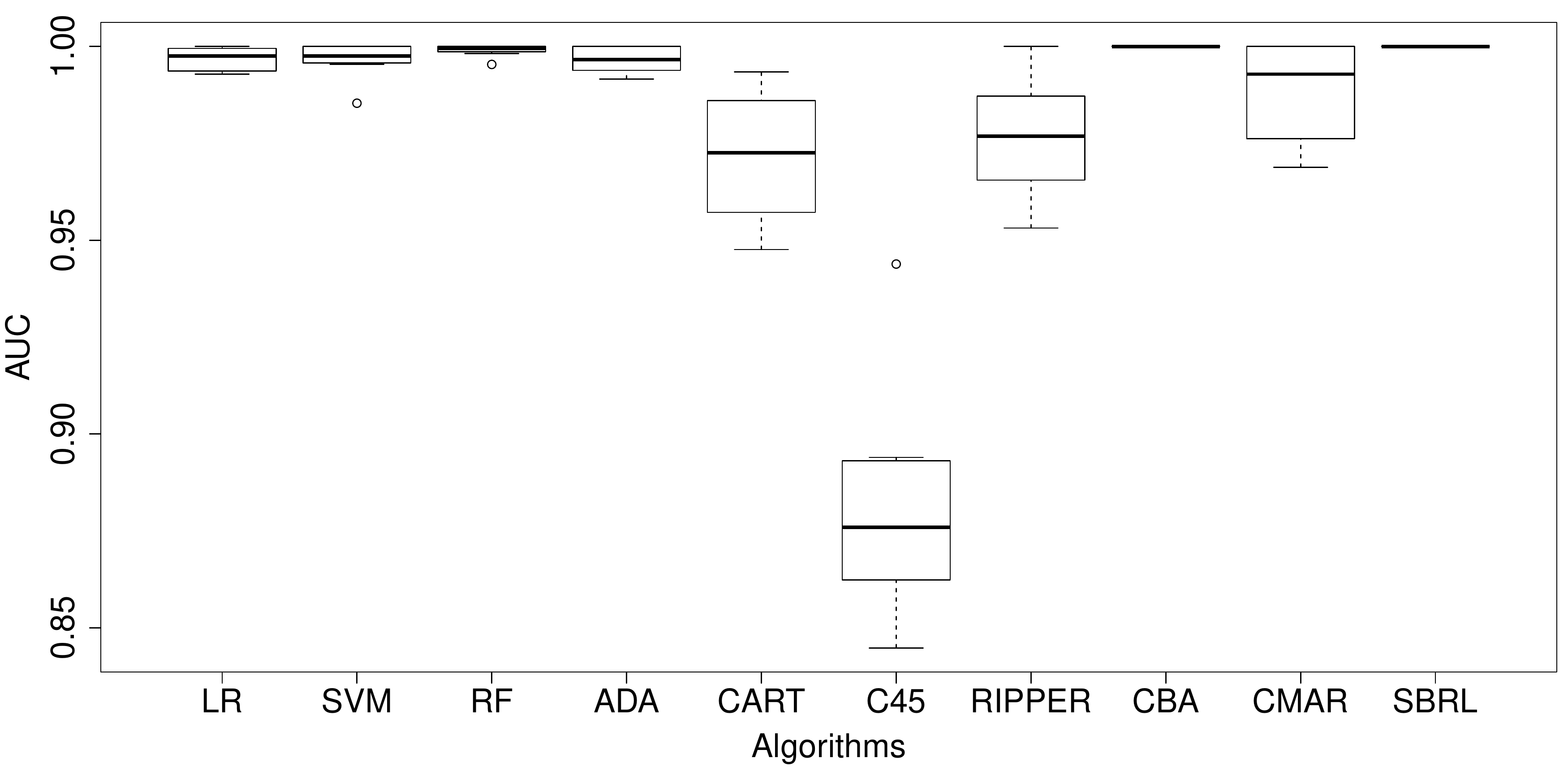}
  \captionof{figure}{Comparison of AUC of ROC among different methods on tic tac toe dataset.}
  \label{figure_auc_comparison_tictactoe}
\end{minipage}%
\begin{minipage}{.5\textwidth}
  \centering
  \includegraphics[width=1.0\linewidth]{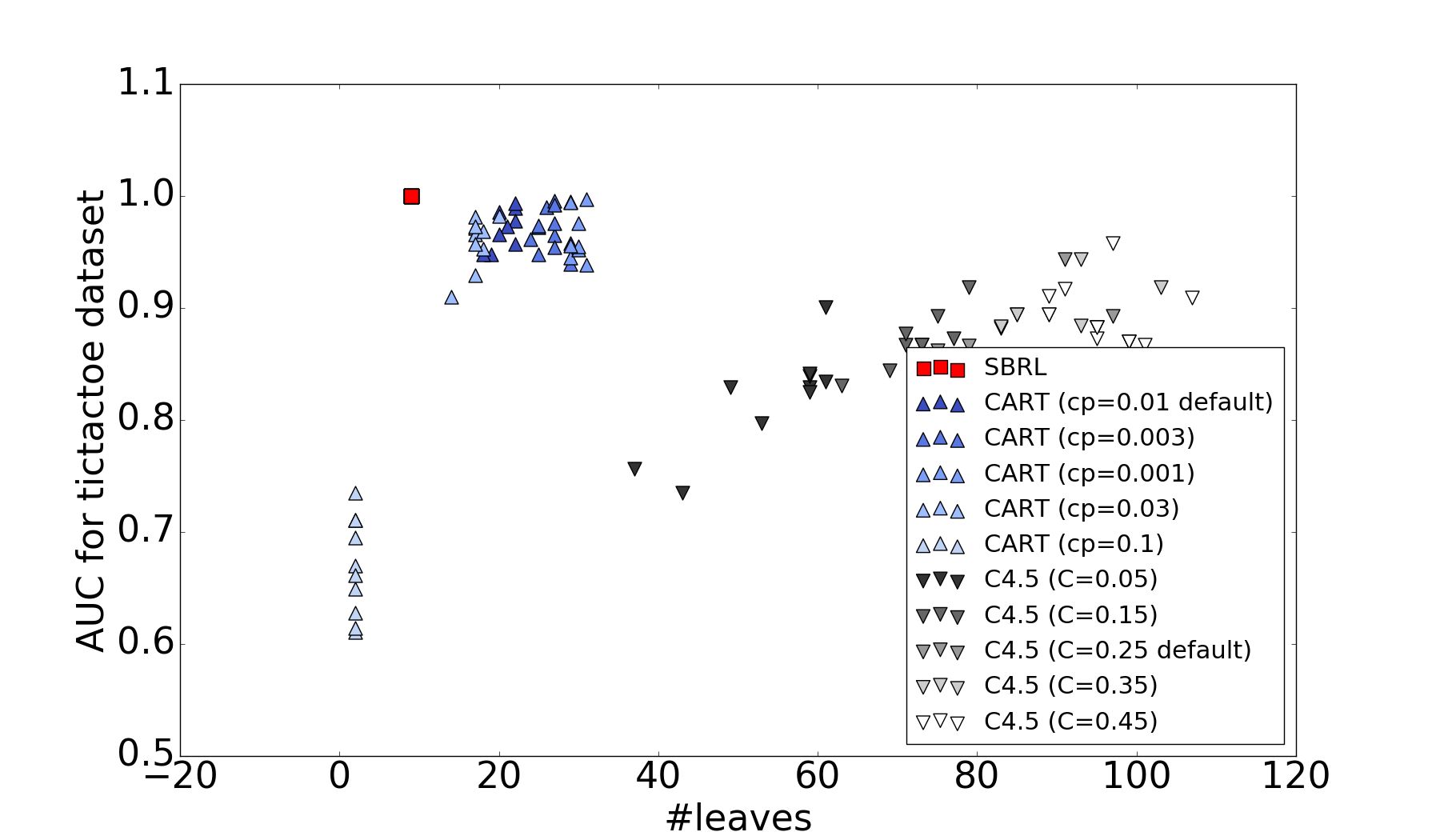}
  \captionof{figure}{Scatter plot of AUC against the number of leaves (sparsity) for the tic tac toe dataset for all 10 folds, for SBRL with one parameter setting, and CART and C4.5 with several parameter settings each.}
  \label{figure_auc_sparsity_tictactoe}
\end{minipage}
\end{figure}

Results for the tic tac toe dataset are shown in Figure \ref{figure_auc_comparison_tictactoe}, Figure \ref{figure_auc_sparsity_tictactoe} and Table \ref{table-runtime-tictactoe}. Each observation in this dataset is a tic tac toe board after the game has finished. If there are 3 X's in a row, the label of the board is 1, otherwise 0. This should not be a difficult learning problem since there are solutions with perfect accuracy on the training set that generalize to the test set. Figure \ref{figure_auc_comparison_tictactoe} shows the sacrifice in AUC made by CART and C4.5. Other methods (RF, SVM, logistic regression, ADA) do not give sparse solutions. In this figure, most of the algorithms were used in their default modes, using their own internal cross-validation routines. The only exceptions are CMAR and CBA, which were tuned to achieve the best accuracy, because their default parameters produce results not comparable to the other methods. This procedure of generating the plots was used for all the figures and tables in this section.

Figure \ref{figure_auc_sparsity_tictactoe} delves further on the decision tree and SBRL models to illustrate the AUC/sparsity tradeoff. It shows a scatter plot of AUC vs$.$ number of leaves, where each point represents an evaluation of one algorithm, on one fold, with one parameter setting. For SBRL, there was no parameter tuning, so there are are 10 points, one for each of the ten folds. We tried many different parameter settings for CART (in blue), and many different parameter settings for C4.5 (in gray), none of which were able to achieve points on the efficient frontier defined by the SBRL method. 

Tables \ref{FigTicTac1}, \ref{FigTicTac2}, and \ref{FigTicTac3} show the models from the the first, second and third SBRL folds.
SBRL's run time was three quarters of a second on average.

\begin{table}
\begin{center}
\resizebox{\textwidth}{!}{\begin{tabular}{| l | l | l | l | l | l | l | l | l | l | l |}
    \hline
    
    Run Time & LR & SVM  & RF   & ADA & CART & C4.5 & RIPPER & CBA & CMAR  & SBRL \\ \hline
    Mean   & 0.044 & 0.194 & 0.382 & 1.648 & 0.023 & 0.133 & 9.810 & 0.129 & 0.084 & 0.759 \\ \hline
    Median & 0.043 & 0.195 & 0.356 & 1.523 & 0.022 & 0.134 & 9.744 & 0.115 & 0.070 & 0.756  \\ \hline
    STD    & 0.004 & 0.007 & 0.067 & 0.338 & 0.001 & 0.003 & 0.171 & 0.044 & 0.044 & 0.020 \\ \hline
    \end{tabular}}
\end{center}
\caption{Run time on tic tac toe dataset \label{table-runtime-tictactoe} in seconds.}
\end{table}
\begin{table}[!htb]
    \begin{minipage}{.47\linewidth}
      \centering
    \begin{tabular}{ l  l  p{1.3cm}}
    Rule-list & PP & Test  Accuracy \\ \hline
if ( x3\&x7\&x5 ), & 0.98 & 1.00 \\
else if ( x1\&x9\&x5 ), & 0.98 & 1.00 \\
else if ( x8\&x2\&x5 ), & 0.98 & 1.00 \\
else if ( x6\&x3\&x9 ), & 0.98 & 1.00 \\
else if ( x4\&x6\&x5 ), & 0.98 & 1.00 \\
else if ( x2\&x1\&x3 ), & 0.98 & 1.00 \\
else if ( x8\&x7\&x9 ), & 0.98 & 1.00 \\
else if ( x4\&x1\&x7 ), & 0.98 & 1.00 \\
else ( default ), & 0.0044 & 1.00 \\
    \end{tabular}
      \caption{Example of rule list for tic tac toe dataset, fold 1 (CV1). PP: probability that the label is positive}
      \label{FigTicTac1}
    \end{minipage}%
    \begin{minipage}{.48\linewidth}
      \centering
    \begin{tabular}{ l  l  p{1.3cm}}
    Rule-list & PP & Test  Accuracy \\ \hline
if ( o9\&o1\&o5 ), & 0.03 & 1.00 \\
else if ( o7\&o3\&o5 ), & 0.026 & 1.00 \\
else if ( o6\&o9\&o3 ), & 0.037 & 1.00 \\
else if ( o8\&o2\&o5 ), & 0.036 & 1.00 \\
else if ( o4\&o6\&o5 ), & 0.04 & 1.00 \\
else if ( o8\&o9\&o7 ), & 0.04 & 1.00 \\
else if ( o2\&o1\&o3 ), & 0.03 & 1.00 \\
else if ( o4\&o7\&o1 ), & 0.04 & 1.00 \\
else ( default ), & 0.97 & 0.98 \\
    \end{tabular}
      \caption{Example of rule list for tic tac toe dataset, fold 2 (CV2).}
      \label{FigTicTac2}
    \end{minipage}
\end{table}

\begin{table}
\begin{center}
    \begin{tabular}{ l  l  l}
    Rule-list & PP & Test Accuracy \\ \hline
if ( x8\&x9\&x7 ), & 0.98 & 1.00 \\
else if ( x6\&x9\&x3 ), & 0.98 & 1.00 \\
else if ( o4\&o1\&o7 ), & 0.042 & 1.00 \\
else if ( o6\&o3\&o9 ), & 0.043 & 1.00 \\
else if ( x4\&x7\&x1 ), & 0.98 & 1.00 \\
else if ( x2\&x1\&x3 ), & 0.98 &  1.00 \\
else if ( o2\&o3\&o1 ), & 0.050 & 1.00 \\
else if ( o5 ), & 0.0079 & 1.00 \\
else if ( o3\&x7 ), & 0.71 & 1.00 \\
else if ( o8\&o7\&o9 ), & 0.040 & 1.00 \\
else ( default ), & 0.99 & 0.96 \\
    \end{tabular}
\end{center}
\caption{Example of rule list for tic tac toe dataset, fold 3 (CV3).\label{FigTicTac3}}
\end{table}

\begin{figure}
\centering
\begin{minipage}{.5\textwidth}
  \centering
  \includegraphics[width=1.0\linewidth]{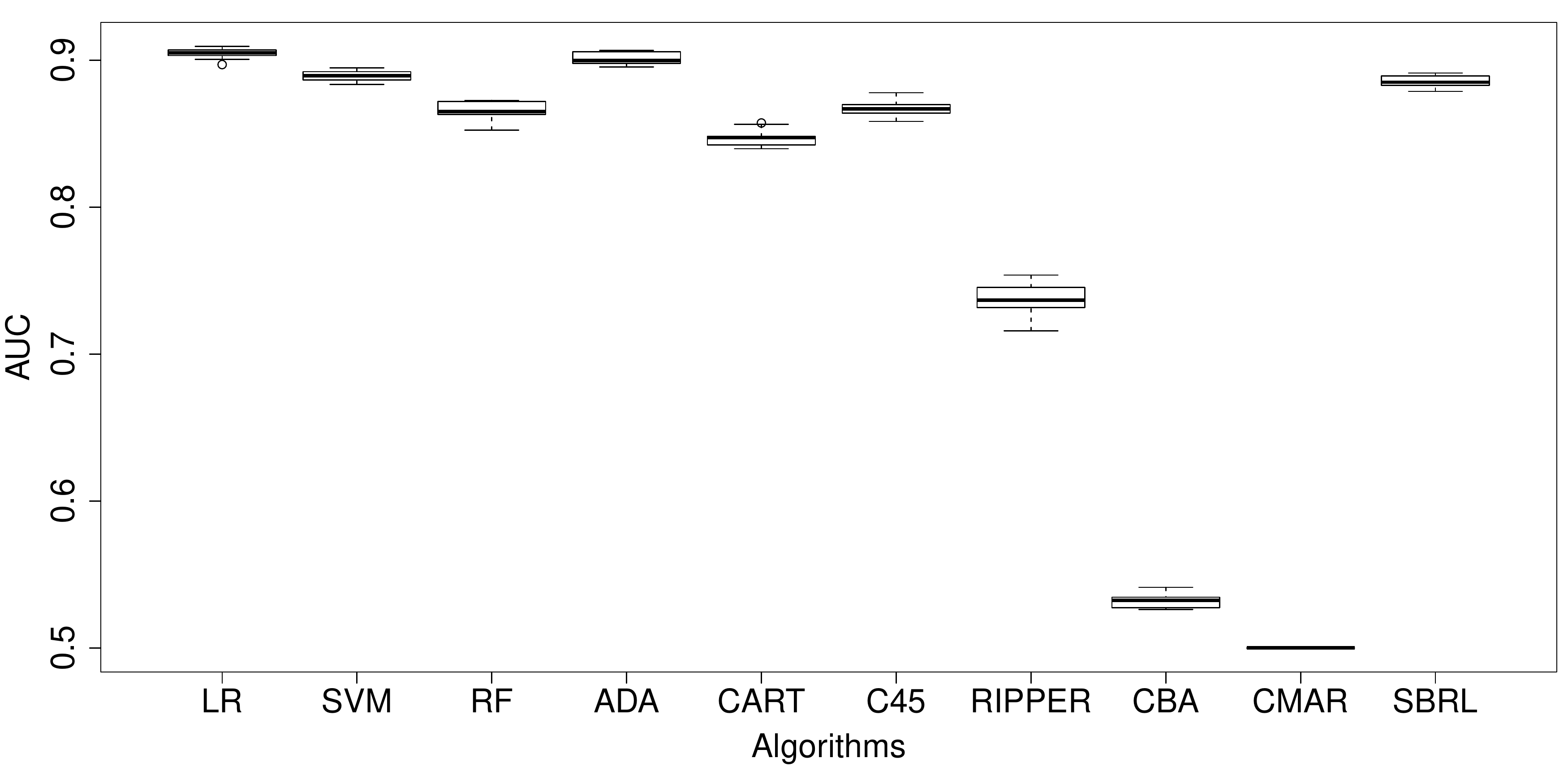}
  \captionof{figure}{Comparison of AUC among different methods on the adult dataset.}
  \label{figure_auc_comparison_adult}
\end{minipage}%
\begin{minipage}{.5\textwidth}
  \centering
  \includegraphics[width=1.0\linewidth]{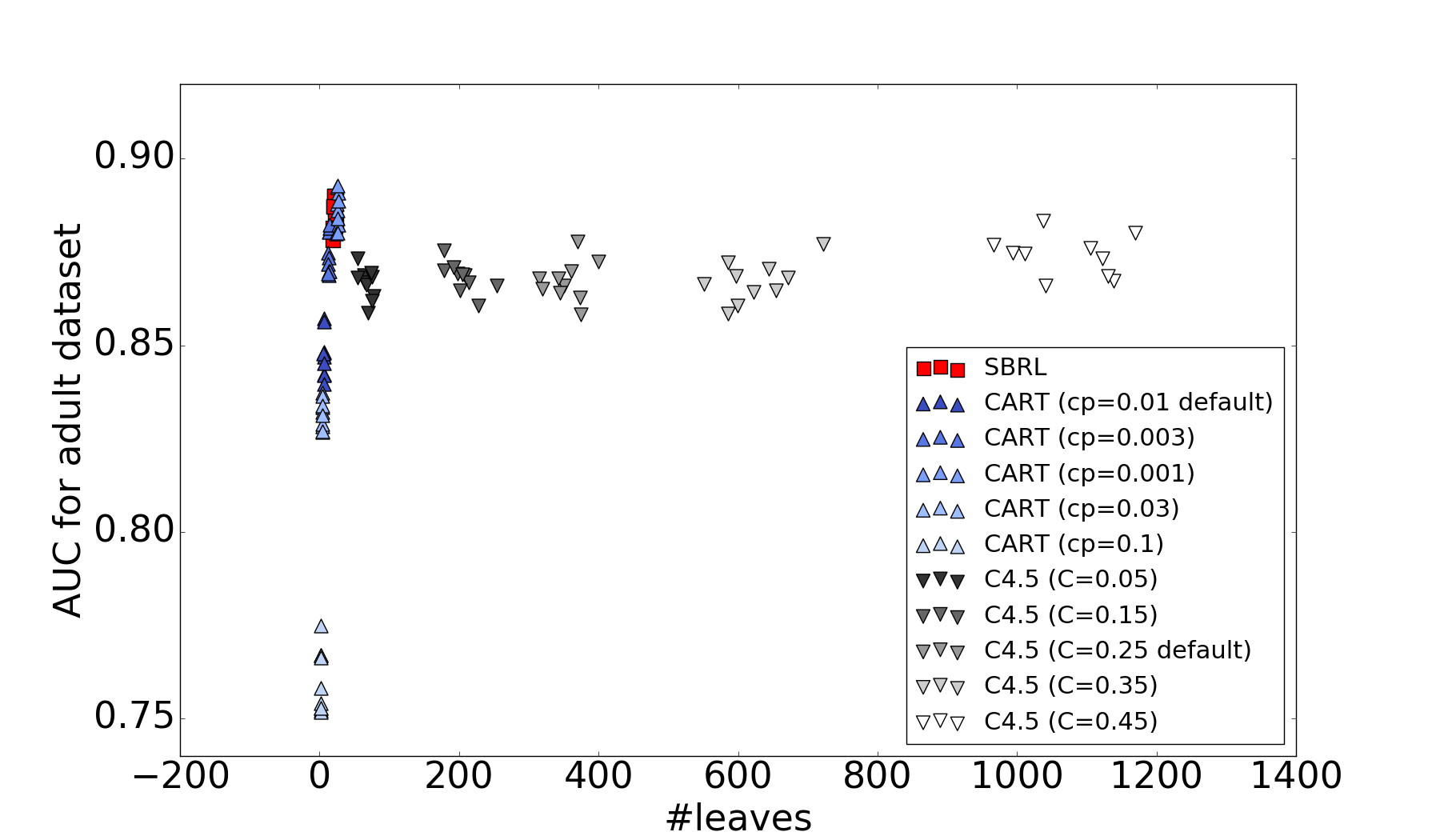}
  \captionof{figure}{Scatter plot of AUC against the number of leaves (sparsity) for the adult dataset for all 10 folds, for SBRL with one parameter setting, and CART and C4.5 with several parameter settings each.}
  \label{figure_auc_sparsity_adult}
\end{minipage}
\end{figure}

\begin{table}[]
\begin{center}
    \resizebox{\textwidth}{!}{\begin{tabular}{| l | l | l | l | l | l | l | l | l | l | l |}
    \hline
    Run Time & LR & SVM  & RF   & ADA & CART & C4.5  & RIPPER & CBA & CMAR & SBRL \\ \hline
    Mean   & 1.353 & 238.5 & 23.53 & 45.71 & 0.809 & 0.512 & 1005.9 & 2.557 & 3.000 & 17.97 \\ \hline
    Median & 1.406 & 239.9 & 23.56 & 43.62 & 0.813 & 0.513 & 1005.4 & 2.520 & 2.920 & 18.01 \\ \hline
    STD    & 0.203 & 5.693 & 0.133 & 4.672 & 0.022 & 0.011 & 100.14 & 0.174 & 0.191 & 0.171 \\ \hline
    \end{tabular}}
\end{center}
\caption{Run time on adult dataset\label{table-runtime-adult} in seconds.}
\end{table}

For the adult dataset, results are in Figure \ref{figure_auc_comparison_adult}, Figure \ref{figure_auc_sparsity_adult} and Table \ref{table-runtime-adult}. The adult dataset contains 45,121 observations and 12 features, where each observation is an individual, and the features are census data, including demographics, income levels, and other financial information. Here, SBRL, which was untuned and forced to be sparse, performed only slightly worse than several of the uninterpretable methods. Its AUC performance dominated those of the CART and C4.5 algorithms. As the scatter plot shows, even if CART were tuned on the test set, it would have performed at around the same level, perhaps slightly worse than SBRL. The timing for SBRL was competitive, at around 18 seconds, where 14 seconds were MCMC iterations. Figure \ref{FigAdultExample} contains one of the rule lists we produced.

\begin{figure}
\centering
\begin{minipage}{.5\textwidth}
  \centering
  \includegraphics[width=1.0\linewidth]{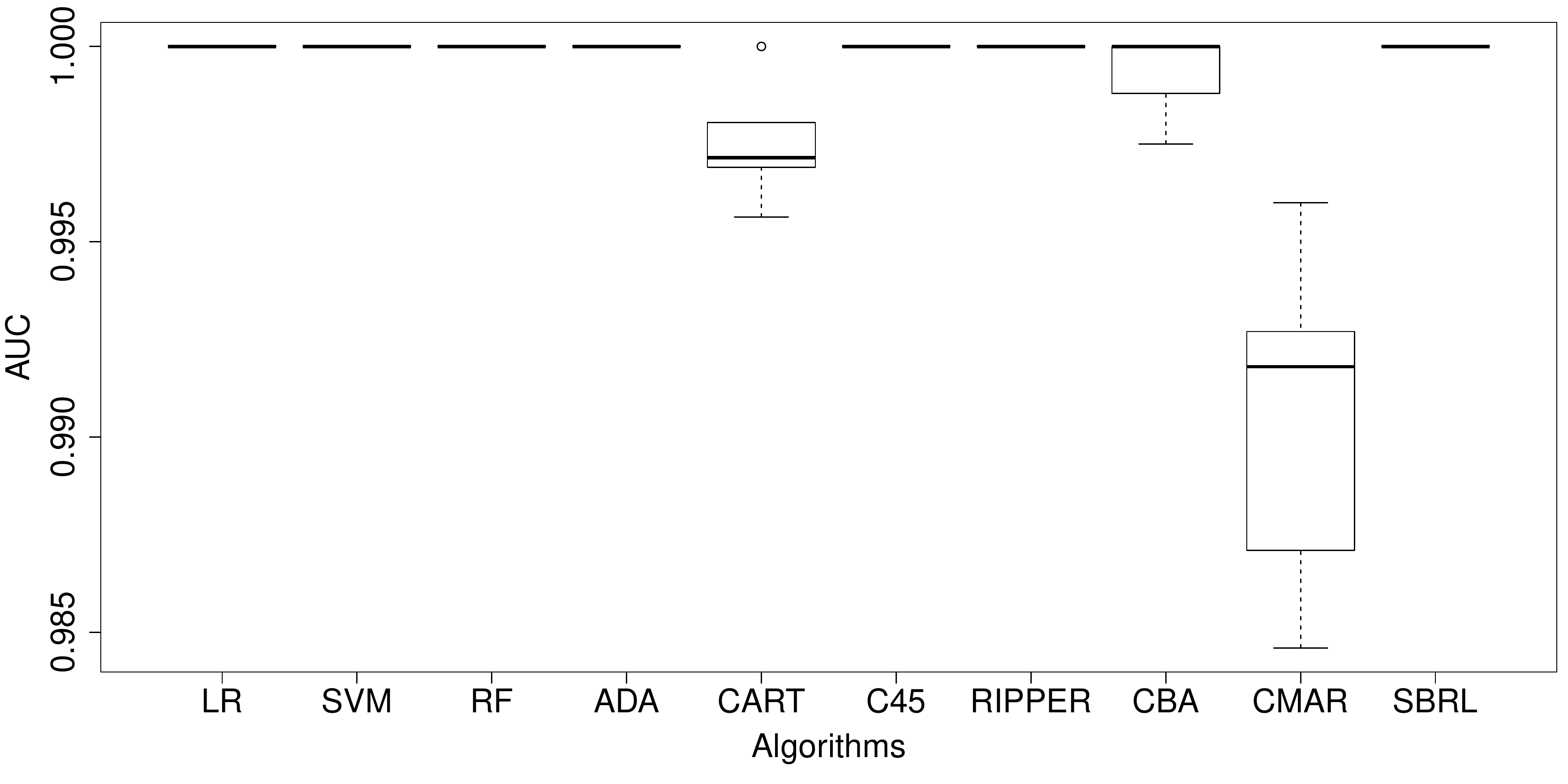}
  \captionof{figure}{Comparison of AUC of ROC among different methods on mushroom dataset.}
 \label{figure_auc_comparison_mushroom}
\end{minipage}%
\begin{minipage}{.5\textwidth}
  \centering
  \includegraphics[width=1.0\linewidth]{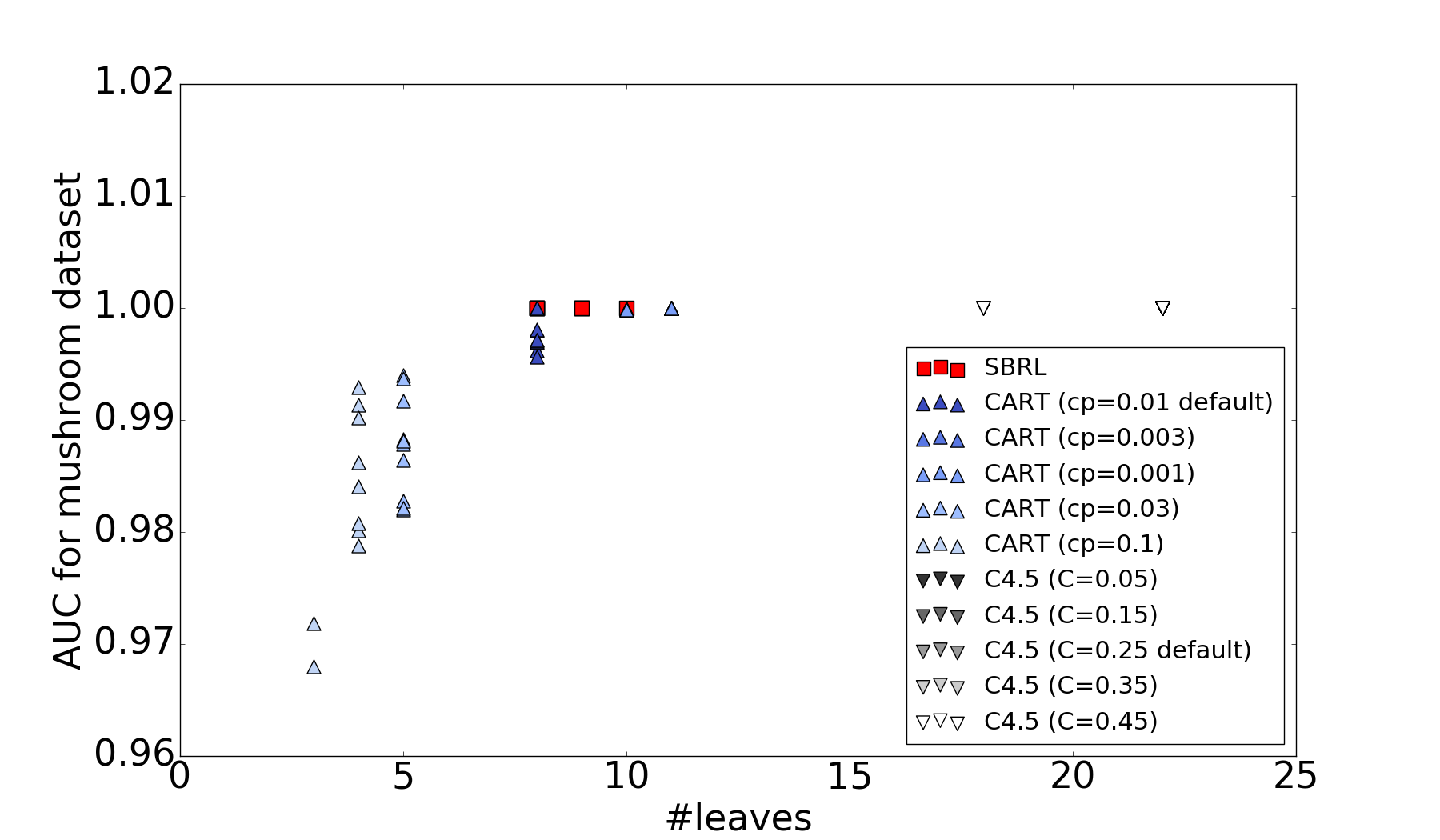}
  \captionof{figure}{Scatter plot of AUC against the number of leaves (sparsity) for the mushroom dataset for all 10 folds, for SBRL with one parameter setting, and CART and C4.5 with several parameter settings each.}
  \label{figure_auc_sparsity_mushroom}
\end{minipage}
\end{figure}

\begin{table}
\begin{center}
    \resizebox{\textwidth}{!}{\begin{tabular}{| l | l | l | l | l | l | l | l | l | l | l |}
    \hline
    Run Time & LR & SVM  & RF   & ADA  & CART & C4.5 & RIPPER & CBA & CMAR & SBRL \\ \hline
    Mean   & 0.791 & 2.691 & 1.922 & 8.001 & 0.132 & 0.257 & 30.26 & 16.48 & 21.25 & 9.196 \\ \hline
    Median & 0.787 & 2.736 & 1.988 & 7.851 & 0.134 & 0.260 & 30.45 & 16.49 & 21.65 & 9.244 \\ \hline
    STD    & 0.026 & 0.128 & 0.235 & 0.354 & 0.010 & 0.019 & 0.690 & 0.568 & 1.027 & 0.208 \\ \hline
    \end{tabular}}
\end{center}
\caption{Run time on mushroom dataset\label{table-runtime-mushroom} in seconds.}
\end{table}

For the mushroom dataset, results are shown in Figure \ref{figure_auc_comparison_mushroom}, Figure \ref{figure_auc_sparsity_mushroom} and Table \ref{table-runtime-mushroom}. Perfect AUC scores were obtained using most of the methods we tried, with the exception of untuned CART, CBA and CMAR. On the scatterplot within Figure \ref{figure_auc_sparsity_mushroom}, there are several solutions with perfect accuracy found by SBRL, tuned CART and and C4.5, of sizes between 8 and 22 rules. The difference in posterior values between the perfect solutions of similar size was extremely small because of our choice of (untuned) $\lambda$. (Tuning and smaller choices for $\lambda$ would improve computation.) Figure \ref{FigMushroomExample} contains one of the rule lists we produced. The CART tree and rule lists produced for this dataset look entirely different. This is because SBRL views each categorical feature as a separate binary variable, whereas CART does not; it can split arbitrarily on categorical variables without penalty. If we had done additional preprocessing on the features to create more splits, we could potentially get rule lists that look like CART's tree. What we got were totally different, yet almost equally perfect, solutions.

\begin{figure}
\centering
\begin{minipage}{.5\textwidth}
  \centering
  \includegraphics[width=1.0\linewidth]{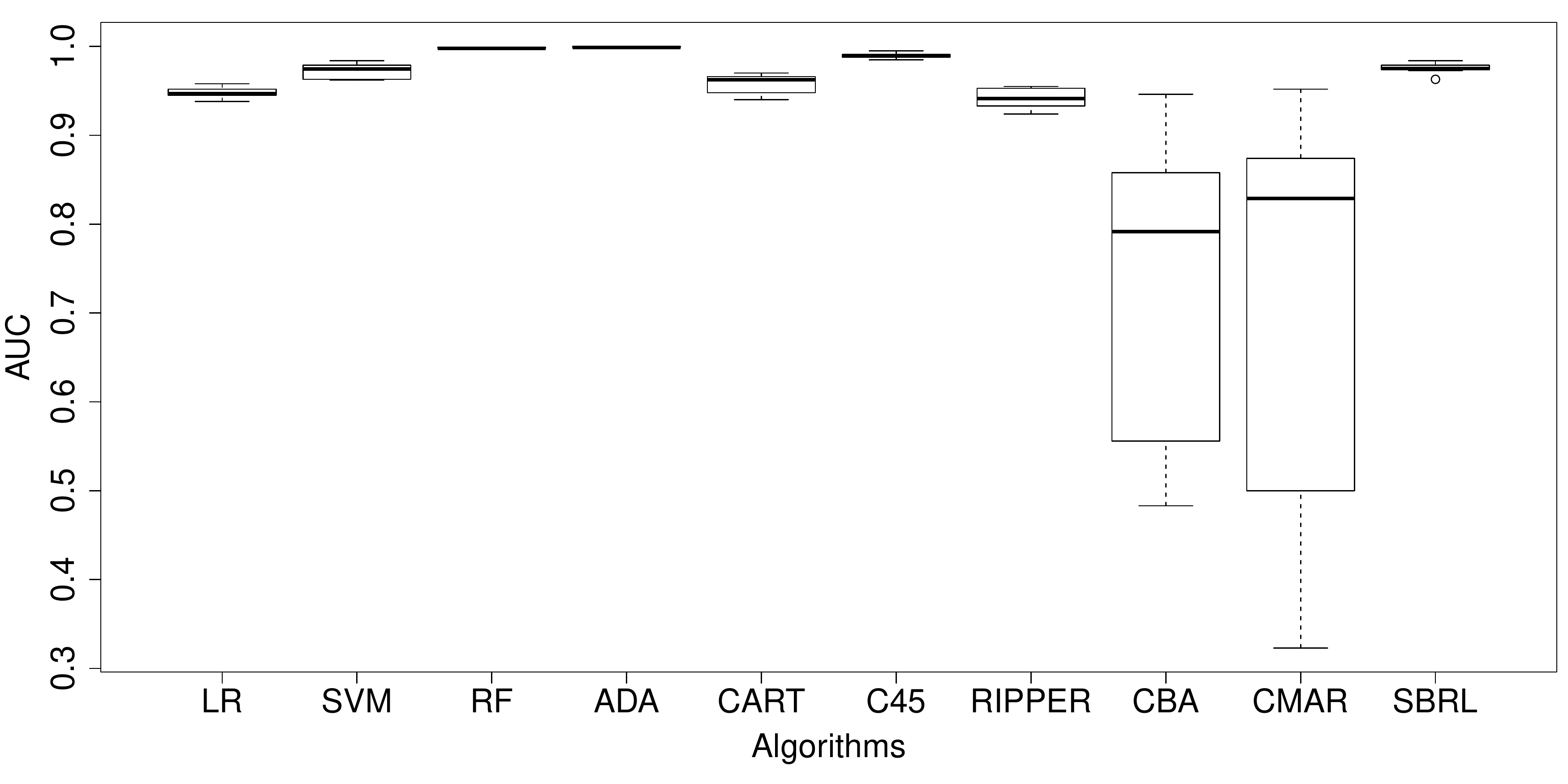}
  \captionof{figure}{Comparison of AUC of ROC among different methods on nursery dataset.}
  \label{figure_auc_comparison_nursery}
\end{minipage}%
\begin{minipage}{.5\textwidth}
  \centering
  \includegraphics[width=1.0\linewidth]{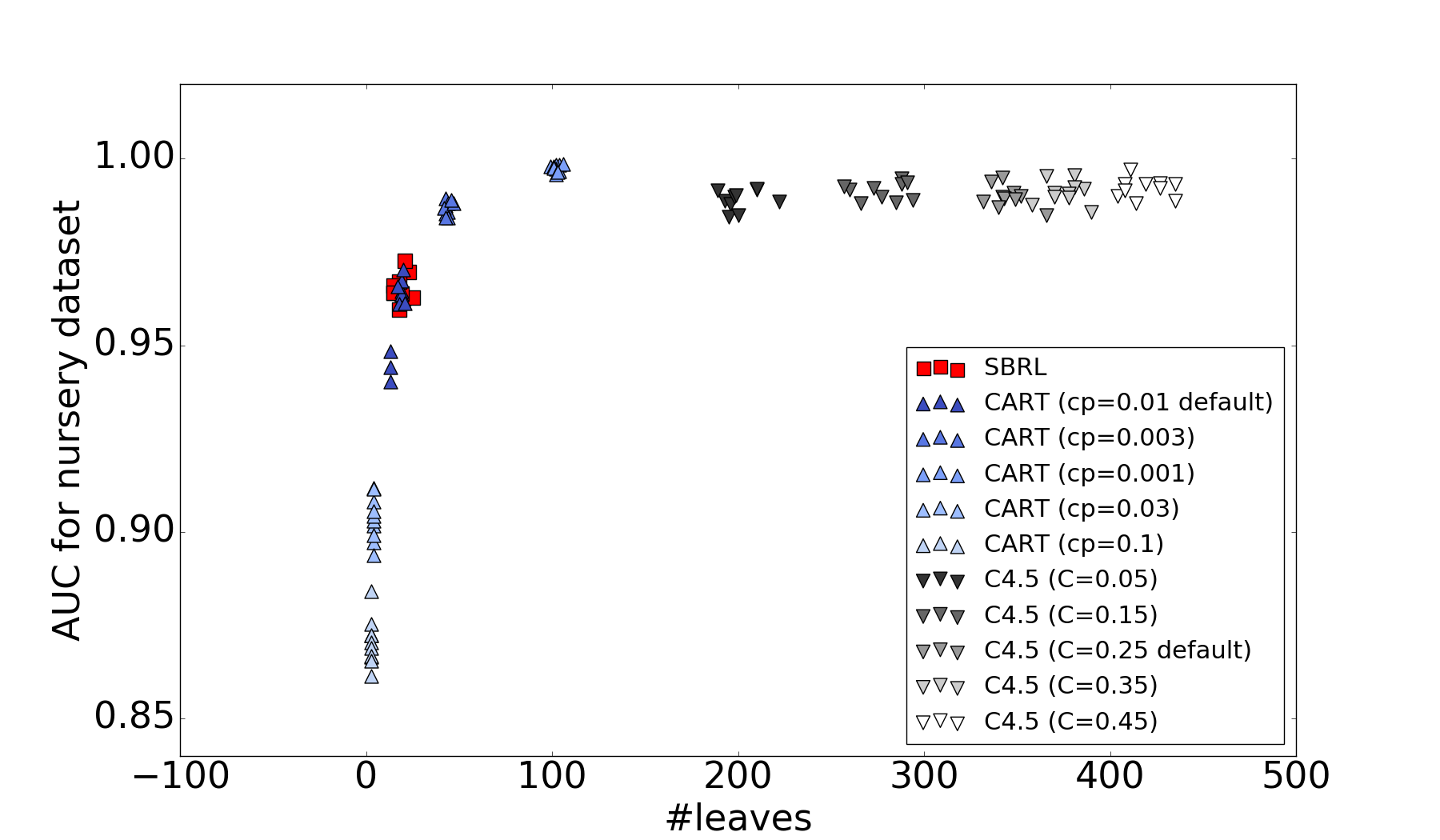}
  \captionof{figure}{Scatter plot of AUC against the number of leaves (sparsity) for the nursery dataset for all 10 folds, for SBRL with one parameter setting, and CART and C4.5 with several parameter settings each.}
  \label{figure_auc_sparsity_nursery}
\end{minipage}
\end{figure}

The results from the nursery dataset are shown in Figure \ref{figure_auc_comparison_nursery}, Figure \ref{figure_auc_sparsity_nursery} and Table \ref{table-runtime-nursery}. A similar story holds as for the previous datasets: SBRL is on the optimal frontier of accuracy/sparsity without tuning and with reasonable run time.

\begin{table}[]
\begin{center}
    \resizebox{\textwidth}{!}{\begin{tabular}{| l | l | l | l | l | l | l | l | l | l | l |}
    \hline
    Run Time & LR & SVM  & RF   & ADA & CART & C4.5  & RIPPER & CBA & CMAR & SBRL \\ \hline
    Mean   & 0.359 & 10.65 & 3.511 & 7.709 & 0.099 & 0.174 & 380.8 & 0.212 & 0.432 & 3.751 \\ \hline
    Median & 0.363 & 10.70 & 3.583 & 7.755 & 0.101 & 0.206 & 380.4 & 0.120 & 0.360 & 3.806 \\ \hline
    STD    & 0.017 & 0.269 & 0.208 & 0.146 & 0.004 & 0.094 & 37.89 & 0.245 & 0.231 & 0.222 \\ \hline
    \end{tabular}}
\end{center}
\caption{Run time of nursery dataset\label{table-runtime-nursery} in seconds.}
\end{table}

\begin{figure}
\centering
\begin{minipage}{.5\textwidth}
  \centering
  \includegraphics[width=1.0\linewidth]{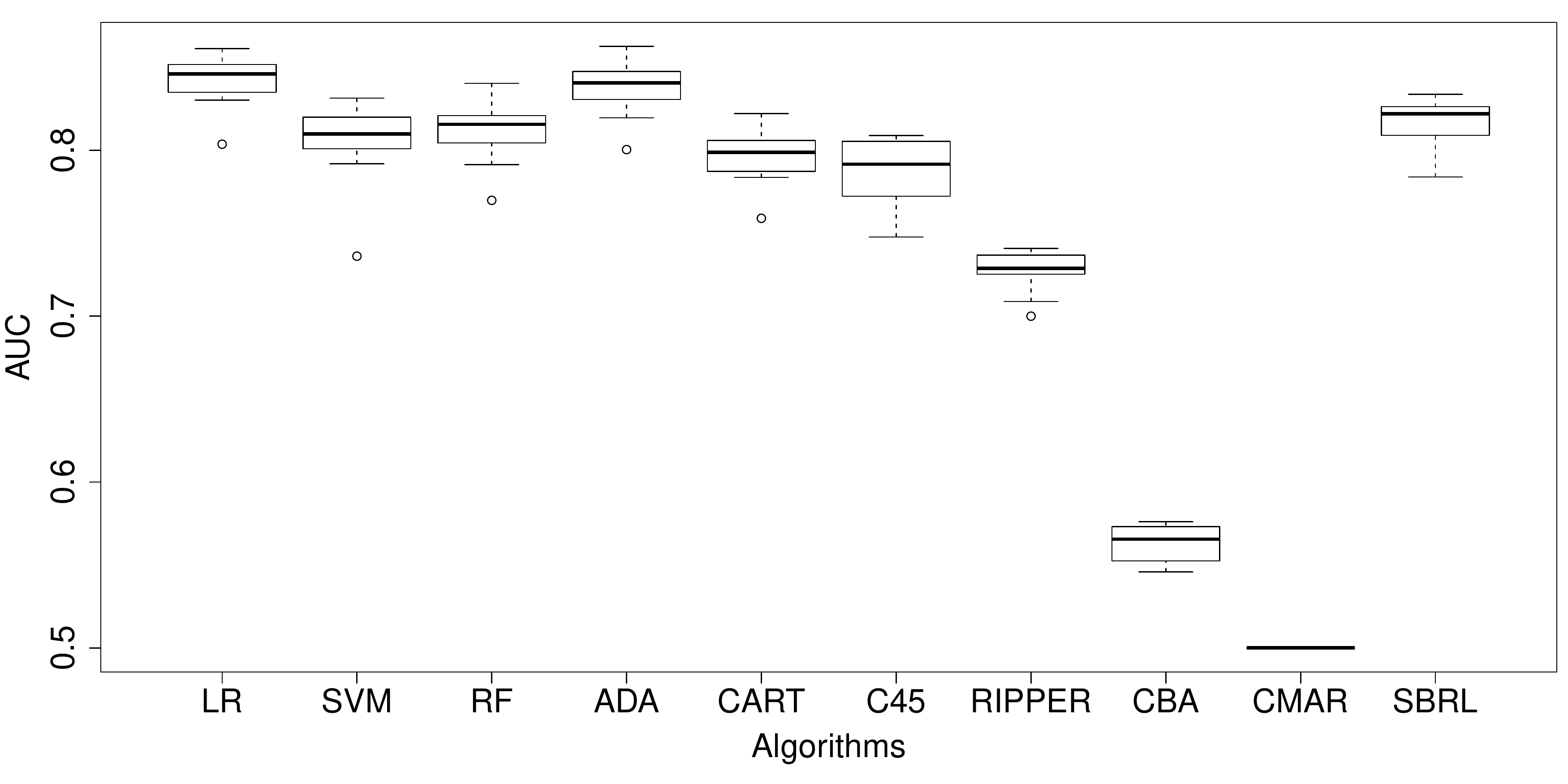}
  \captionof{figure}{Comparison of AUC of ROC among different methods on telco dataset.}
  \label{figure_auc_comparison_telco}
\end{minipage}%
\begin{minipage}{.5\textwidth}
  \centering
  \includegraphics[width=1.0\linewidth]{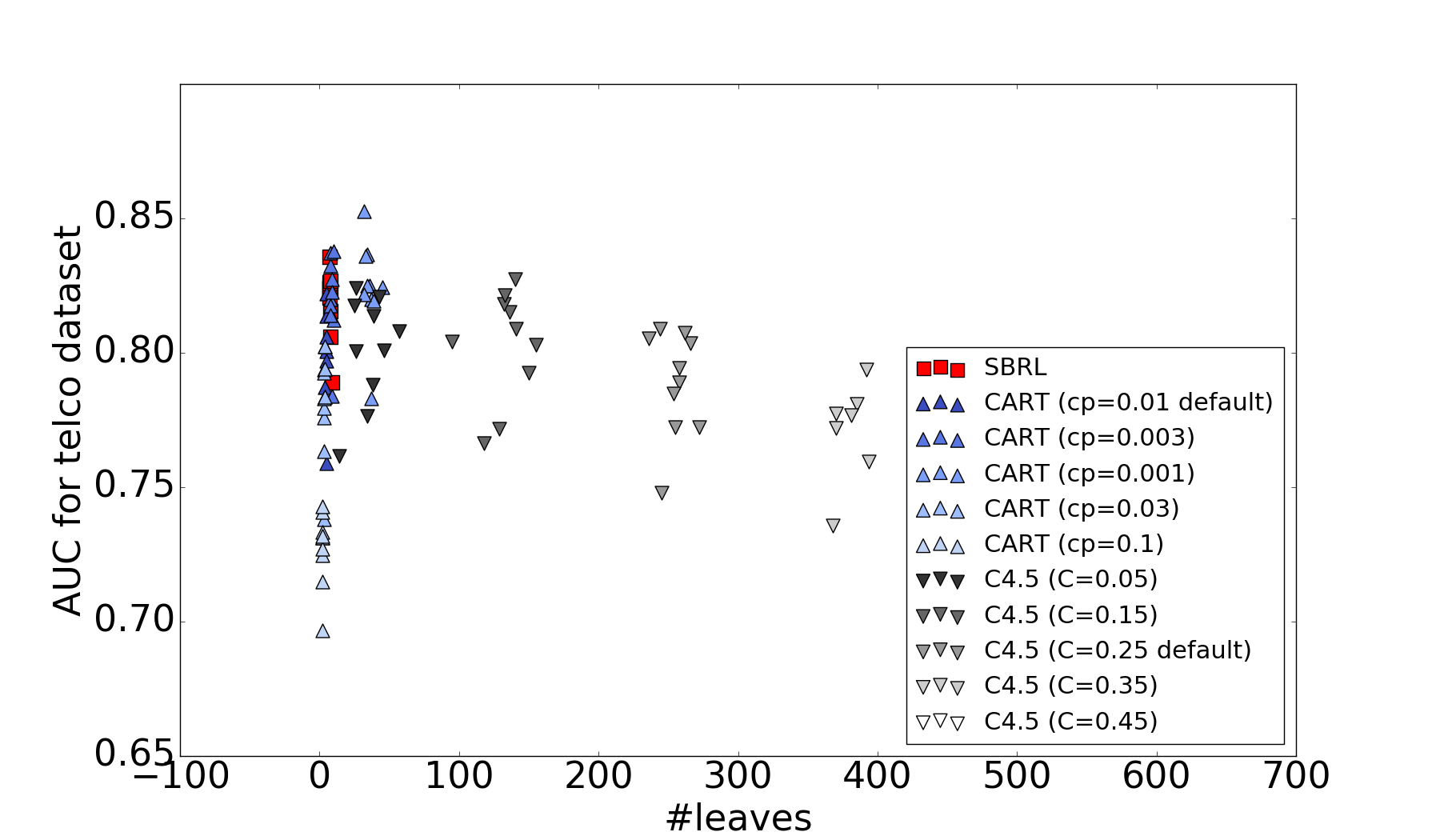}
  \captionof{figure}{Scatter plot of AUC against the number of leaves (sparsity) for the telco dataset for all 10 folds, for SBRL with one parameter setting, and CART and C4.5 with several parameter settings each.}
  \label{figure_auc_sparsity_telco}
\end{minipage}
\end{figure}

Figure \ref{figure_auc_comparison_telco}, Figure \ref{figure_auc_sparsity_telco} and Table \ref{table-runtime-telco} show the results for the telco dataset, which contains 7043 observations and 18 features. Similar observations hold for this dataset. The models from three of the ten folds are provided in Tables \ref{FigTelco1}, \ref{FigTelco2} and  \ref{FigTelco3}. These models illustrate that generally, rule lists are not the same between folds, but often tend to use similar rules.


\begin{table}
\begin{center}
    \resizebox{\textwidth}{!}{\begin{tabular}{| l | l | l | l | l | l | l | l | l | l | l |}
    \hline
    Run Time & LR & SVM & RF   & ADA  & CART & C4.5  & RIPPER & CBA & CMAR & SBRL \\ \hline
    Mean   & 0.267 & 7.468 & 3.703 & 7.839 & 0.168 & 0.250 & 37.14 & 8.028 & 1.679 & 5.239 \\ \hline
    Median & 0.272 & 7.550 & 3.695 & 8.726 & 0.168 & 0.252 & 37.63 & 8.050 & 1.705 & 5.271 \\ \hline
    STD    & 0.009 & 0.207 & 0.183 & 0.111 & 0.008 & 0.017 & 3.202 & 0.400 & 0.161 & 0.149 \\ \hline
    \end{tabular}}
\end{center}
\caption{Run time of telco dataset\label{table-runtime-telco} in seconds.}
\end{table}


\begin{table}
\begin{center}
    \begin{tabular}{l l l}
    Rule-list & PP & Test Accuracy \\ \hline
if ( Contract=One\_year \&StreamingMovies=Yes ), & 0.20 & 0.82 \\
else if ( Contract=One\_year ), & 0.050 & 0.96 \\
else if ( tenure$<$1year \&InternetService=Fiber\_optic ), & 0.70 & 0.71 \\
else if ( Contract=Two\_year ), & 0.029 & 0.97 \\
else if ( InternetService=Fiber\_optic \&OnlineSecurity=No ), & 0.48 & 0.58 \\
else if ( OnlineBackup=No \&TechSupport=No ), & 0.41 & 0.61 \\
else ( default ), & 0.22 & 0.78 \\
    \end{tabular}
\end{center}
      \caption{Example of rule list for telco dataset fold 1 (CV1).\label{FigTelco1}}
\end{table}

\begin{table}
\begin{center}
    \begin{tabular}{l l l}
    Rule-list & PP & Test Accuracy \\ \hline
if ( Contract=One\_year \&StreamingMovies=Yes ), & 0.20 & 0.81 \\
else if ( tenure$<$1year \&InternetService=Fiber\_optic ), & 0.70 & 0.70 \\
else if ( tenure$<$1year \&OnlineBackup=No ), & 0.44 & 0.57 \\
else if ( InternetService=Fiber\_optic\\ $\quad\quad$\&Contract=Month-to-month ), & 0.43 & 0.57 \\
else if ( Contract=Month-to-month ), & 0.22 & 0.82 \\
else ( default ), & 0.034 & 0.97 \\
    \end{tabular}
      \caption{Example of rule list for telco dataset fold 2 (CV2).\label{FigTelco2}}
\end{center}
\end{table}

\begin{table}
\begin{center}
    \begin{tabular}{ l  l  l}
    Rule-list & PP & Test Accuracy \\ \hline
    if ( Contract=One\_year\&StreamingMovies=Yes ), & 0.20 & 0.81 \\
else if ( Contract=Two\_year ), & 0.032 & 0.98 \\
else if ( Contract=One\_year ), & 0.054 & 0.97 \\
else if ( tenure$<$1year\&InternetService=Fiber\_optic ), & 0.70 & 0.72 \\
else if ( PaymentMethod=Electronic\_check\\ $\quad\quad$\&InternetService=Fiber\_optic ), & 0.48 & 0.45 \\
else ( TechSupport=No\&OnlineSecurity=No ), & 0.42 & 0.64 \\
else ( default ), & 0.22 & 0.78 \\
    \end{tabular}
\end{center}
\caption{Example of rule list for telco dataset CV3.\label{FigTelco3}}
\end{table}

\begin{figure}
\centering
\begin{minipage}{.5\textwidth}
  \centering
  \includegraphics[width=1.0\linewidth]{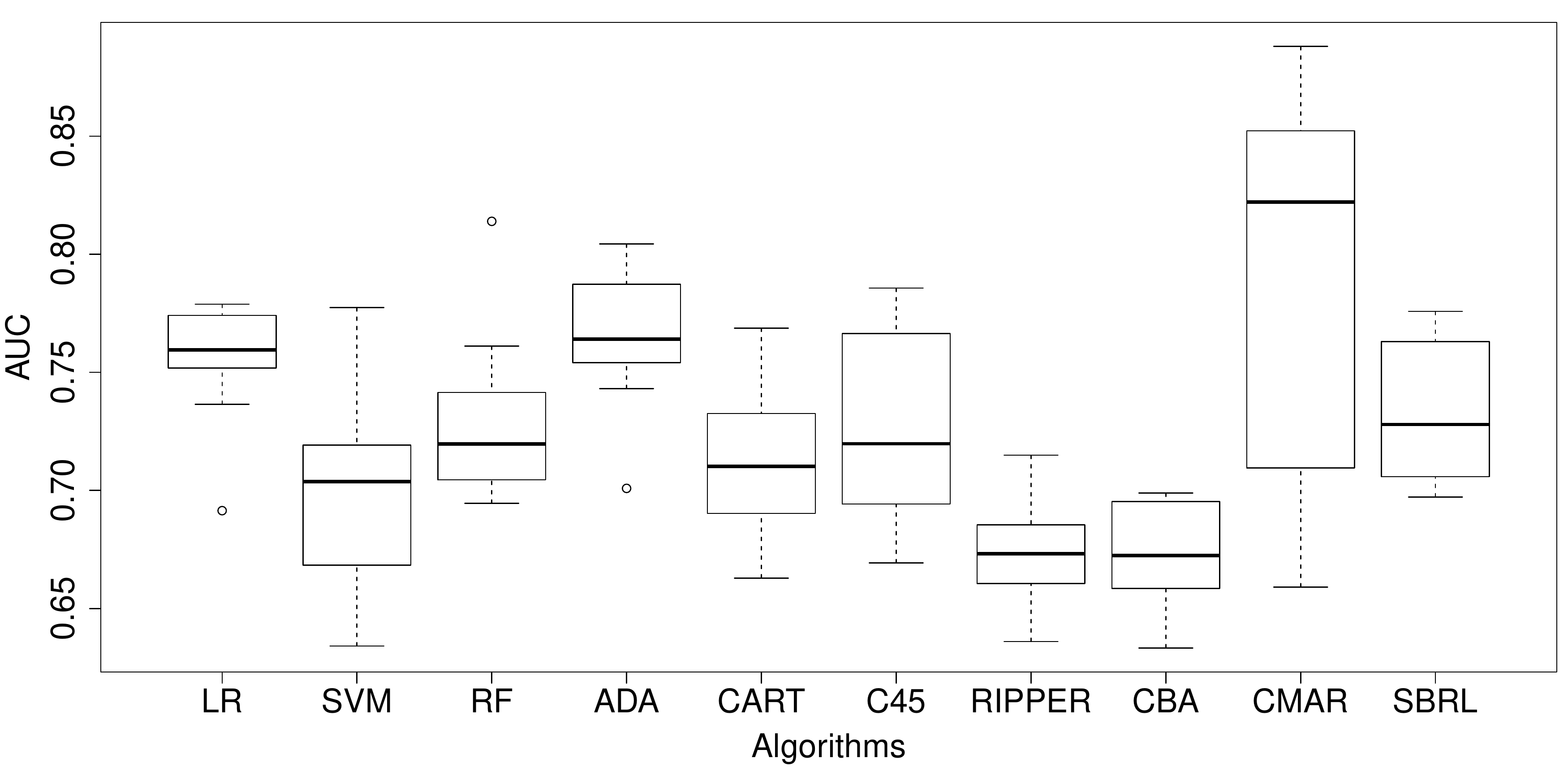}
  \captionof{figure}{Comparison of AUC of ROC among different methods on titanic dataset.}
  \label{figure_auc_comparison_titanic}
\end{minipage}%
\begin{minipage}{.5\textwidth}
  \centering
  \includegraphics[width=1.0\linewidth]{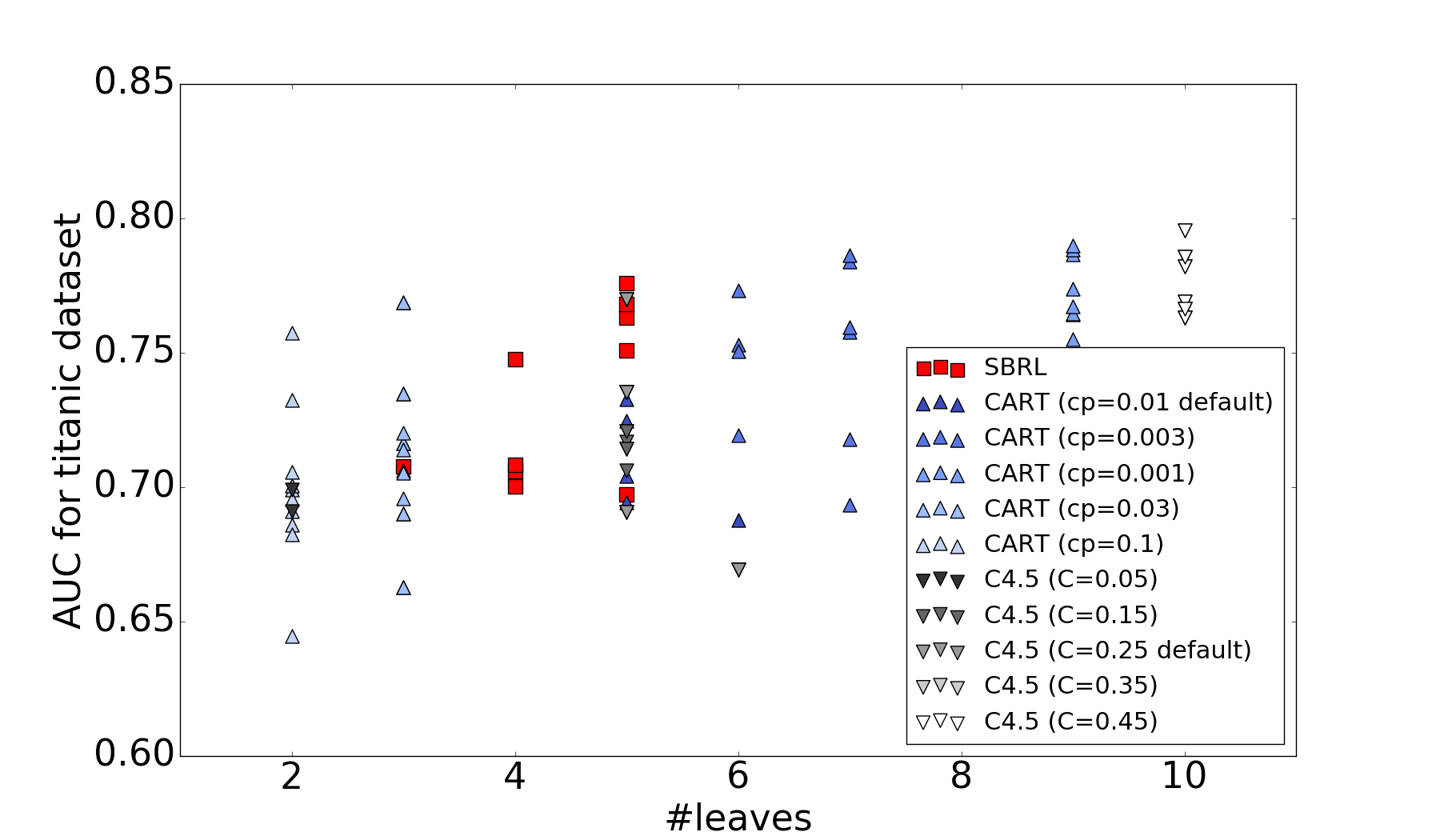}
  \captionof{figure}{Scatter plot of AUC against the number of leaves (sparsity) for the titanic dataset for all 10 folds, for SBRL with one parameter setting, and CART and C4.5 with several parameter settings each.}
  \label{figure_auc_sparsity_titanic}
\end{minipage}
\end{figure}

The titanic dataset evaluation results are in Figures \ref{figure_auc_comparison_titanic}, \ref{figure_auc_sparsity_titanic} and Table \ref{table-runtime-titanic}. The titanic dataset contains data about 2201 passengers and crew aboard the Titanic. 

\begin{table}
\begin{center}
    \resizebox{\textwidth}{!}{\begin{tabular}{| l | l | l | l | l | l | l | l | l | l | l |}
    \hline
    Run Time & LR & SVM & RF   & ADA  & CART & C4.5  & RIPPER & CBA & CMAR & SBRL \\ \hline
    Mean   & 0.015 & 0.330 & 0.470 & 1.343  & 0.016 & 0.089 & 6.257 & 0.005 & 0.004 & 0.357 \\ \hline
    Median & 0.015 & 0.332 & 0.445 & 1.301  & 0.018 & 0.088 & 6.277 & 0.000 & 0.000 & 0.359 \\ \hline
    STD    & 0.002 & 0.011 & 0.108 & 0.145  & 0.007 & 0.007 & 0.072 & 0.009 & 0.009 & 0.011 \\ \hline
    \end{tabular}}
\end{center}
\caption{Run time of titanic dataset\label{table-runtime-titanic} in seconds.}
\end{table}

The results on all of these datasets are consistent. On each dataset, SBRL produces results that are reliable (unlike CART) and sparse (unlike C4.5). The run times are longer but still reasonable and also adjustable since the user can pre-determine exactly how long to run the method. 

\section{Scalability}
We wanted to see how well SBRL could handle larger datasets. We used 1 million datapoints from the USCensus1990 dataset \cite<see>{Bache+Lichman:2013} and set SBRL's parameter to extract $\approx$1 thousand rules as problem (A), and about 50 thousand data points with 50 thousand rules as problem (B). The runtime comparison with CART is shown in Table \ref{table-runtime-USCensus1990}. For problem (A) the run times are similar. For (B) SBRL is slower (2.5 hours) which is not prohibitive for important problems; here one can see why CART does not perform well in high dimensions, as it often spends less time on harder problems than is required to achieve better performance.

\begin{table}
\begin{center}
    {\begin{tabular}{| l | l | l |}
    \hline
    Run Time(s) & SBRL & CART\\ \hline
    (A) & 2700 & 2000 \\ \hline
    (B) & 9000 & 84 \\ \hline
    \end{tabular}}
\end{center}
\caption{Run time of SBRL on USCensus1990 dataset in seconds. A: 1 million data points and 1 thousand rules; B: 50 thousand data points and 50 thousand rules \label{table-runtime-USCensus1990}}
\end{table}

\begin{figure}
\centering
\begin{minipage}{.4\textwidth}
  \centering
  \includegraphics[width=1.0\linewidth]{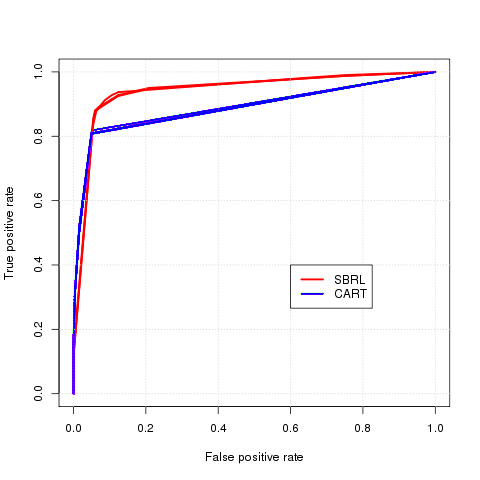}
  \captionof{figure}{AUC plot of SBRL and CART on USCensus1990 for problem A.}
  \label{figure_ROC_a}
\end{minipage}%
\begin{minipage}{.4\textwidth}
  \centering
  \includegraphics[width=1.0\linewidth]{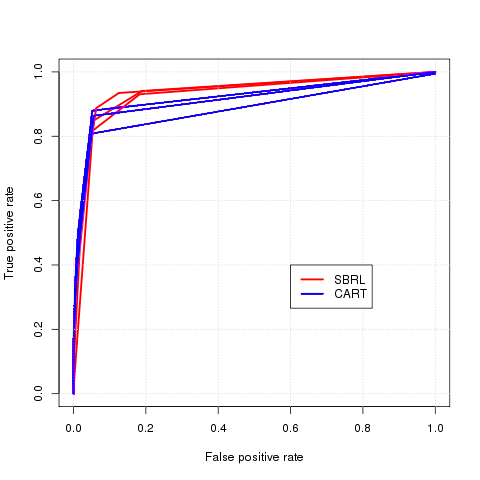}
  \captionof{figure}{AUC plot of SBRL and CART on USCensus1990 for problem B.}
  \label{figure_ROC_b}
\end{minipage}
\end{figure}

\section{Related Works and Discussion}
\label{SecRelated}
Rule learning algorithms have been developed possibly since the AQ algorithm in the 1960s \cite{Michalski1969b} and throughout the 1980's and 90's \cite{Cendrowska1987,Quinlan1983,Clark1989,Cohen95fasteffective,Liu98}. 

Rule lists are not very different from decision trees in capacity; any decision tree can be made into a decision list simply by placing a rule in the list to represent each leaf of the decision tree. Rule lists are automatically a type of decision tree. Thus this method is really a direct competitor for CART.

Interpretability has long since been a fundamental topic in artificial intelligence \cite<see>{ruping2006learning,bratko1997machine,dawes1979robust,VellidoEtAl12,Giraud98,Holte93,Schmueli10,Huysmans11,Freitas14}. Because the rule lists created by our method are designed to be interpretable, one would probably not want to boost them, or combine them in other ways to form more complicated models. This contrasts with, for instance, \citeA{Friedman08}, who linearly combine pre-mined rules.

This work enables us to globally control decision trees in a sense, which could lead to more interesting styles of trees, and different forms of interpretability. For example, one cannot easily construct a Falling Rule List with a greedy splitting method, but can construct one with a global optimization approach. A Falling Rule List \cite{WangRu15} is a decision list where the probabilities of success decrease as we descend along the list. This means we can target the highest probability subgroup by checking only a few conditions. A Causal Falling Rule List (CFRL) \cite{WangRu15CFRL} is another such example. These model causal effects (conditional differences) rather than outcomes. The first rule in the list pinpoints the subgroup with the largest treatment effect. It is possible that many other exotic types of constrained models could be constructed in a computationally efficient way using the ideas in this paper. One could go beyond logical models and consider also mixed logical/linear models \cite<see>{WangOblique15}.

Rule lists and their variants are currently being used for text processing \cite{King14}, discovering treatment regimes \cite{ZhangEtAl15}, and creating medical risk assessments \cite{LethamRuMcMa15,Souillard15}, among other applications. 

There are other subfields where one would pre-mine rules and use them in a classifier. Inductive logic programming \cite{muggleton1994inductive}, greedy top-down decision list algorithms \cite{Rivest87,Sokolova03,Anthony05,Marchand05,RudinLeMa13,Goessling2015}, associative classification \cite{Vanhoof10,Liu98,Li01,Yin03} and its Bayesian counterparts  \cite{McCormick:2011ws} all fall into this category. None of the methods in these fields follow the same general procedure as we do, where rules are fully optimized into an optimal tree using low-level computations, and where rules are eliminated based on theoretical motivation, as we have in Sections \ref{SecTheor}.

Teleo-reactive programs \cite{Nilsson94} use a decision list structure and could benefit from learning this structure from data.

There are a series of works from the mid-1990's on finding optimal decision trees using dynamic programming and search techniques \cite<e.g.,>{Bennett96optimaldecision,Auer95theoryand,dobkininduction}, mainly working with only fixed depth trees. None of these works use the systems level techniques we use to speed up computation. \citeA{FarhangfarGZ08} use a screening step that reduces the number of features, using the extremly strong Na\"ive Bayes assumption that the features are independent, given the class, and then uses dynamic programming to construct an optimal fixed-depth tree. One particularly interesting work following this literature is that of \citeA{NijssenFromont2010}, which allows for pre-mined rules to form trees, but in a different way than our method or associative classifiers. \citeA{NijssenFromont2010} has the user  pre-mine all possible \textit{leaves}, enumerating all conditions leading to that leaf. (By contrast, in our work and in associative classification, we mine only small conjunctions, and their ordered combination creates leaves.)  \citeA{NijssenFromont2010} warn about issues related to running out of memory. As a possible extension, the work proposed here could be modified to handle regularized empirical risk minimization, in particular it could use the objective of \citeA{RudinEr15}, which is a balance between accuracy and sparsity of rule lists. It could also be modified to handle disjunctive normal form classifiers, for which there are now Bayesian models analogous to the ones studied in this work \cite{WangEtAl15}. Bayesian tree models may also be able to be constructed using our setup, where one would mine rules and create a globally optimal tree \cite{Dension:1998hl,Chipman:2002hc,Chipman10}. It may be logistically more difficult to code trees than lists in order to take advantage of the fast lower level computations, but this is worth further investigation.


A theoretical result states that the VC (Vapnik-Chervonenkis) dimension of the set of rule lists created using pre-mined rules is exactly the size of the set of pre-mined rules \cite{RudinLeMa13}. This provides a connection to linear models, whose complexity is the number of features plus 1. That is, the VC dimension of rule lists created from $|\mathcal{A}|$ predefined rules is essentially the same as that of linear models with $|\mathcal{A}|$ features. If some rules are eliminated (for instance based on the theorems in Section \ref{SecTheor}) then the VC dimension is the size of the set of rules that remain.

An extension of this work \cite{AngelinoEtAl2017} does not provide probabilistic predictions, but is able to provide a certificate of optimality to a globally optimal rule list. This indicates that SBRL is probably also achieving optimality; however, because SBRL is probabilistic, the proof of optimality is much more difficult. To clarify: finding the optimal solution for both methods should be approximately equally difficult, but proving optimality for SBRL is much more difficult. There is a clear benefit to SBRL's probabilistic predictions however.

\section*{Conclusion}

We finish by stating why/when one would want to use this particular method. SBRL is not meant as a competitor for black box classifiers such as neural networks, support vector machines, gradient boosting or random forests. It is useful when machine learning tools are used as a decision aid to humans, who need to understand the model in order to trust it and make data-driven decisions. SBRL is not a greedy splitting/pruning procedure like decision tree algorithms (CART, C4.5), which means that it more reliably computes high quality solutions, at the possible expense of additional computation time. Many of the decision tree methods do not compute sparse trees and do not provide interpretable models, as we have seen with C4.5. Our code is a strict improvement over the original Bayesian Rule Lists algorithm if one is looking for a maximum a posteriori solution. It is faster because of careful use of low level computations and theoretical bounds.

\acks{The authors would like to acknowledge partial funding provided by  Philips, Wistron, and Siemens.}

\section*{Code}
Code for SBRL is available at the following link: https://github.com/Hongyuy/sbrlmod \\
Link to R package sbrl on CRAN: https://cran.r-project.org/web/packages/sbrl/index.html

\appendix
\section{Proof of Theorem \ref{TheoremUpperBoundForm}}

To prove this, we will show that any rule list with more than $m_{\max}$ rules has a lower posterior than the trivial empty rule list. This means any rule list with more than $m_{\max}$ terms cannot be a MAP rule list. Denote $\phi$ as the trivial rule list with only the default rule. By definition of $d^*$ as a MAP rule list, it has a posterior at least as high as $\phi$.
\begin{eqnarray*}
\lefteqn{\textrm{Posterior}(d^*|\mathcal{A}, X, Y, \alpha, \lambda, \eta) \geq \textrm{Posterior}(\phi|\mathcal{A}, X, Y, \alpha, \lambda, \eta) }\\
\lefteqn{\frac{(\lambda^{m^*}/m^*!)}{\sum\limits_{j=0}^{|\mathcal{A}|} (\lambda^j/j!)}
\prod_{j=0}^{m^*} \left(\frac{(\eta^{c_j}/c_j!)}{\sum\limits_{k \in R_{j-1}(c_{<j},\mathcal{A})}  (\eta^k/k!)} \frac{1}{|Q_{c_j}|}
\frac{ \Gamma (N_{j,0}+\alpha_0)\Gamma (N_{j,1}+\alpha_1)}{ \Gamma(N_{j,0} +N_{j,1}+\alpha_0+\alpha_1)}\right)
}\\
&\geq& 
\frac{\lambda^0/0!}{\sum\limits_{j=0}^{|\mathcal{A}|} (\lambda^j/j!)} 
\frac{\Gamma(N_-+\alpha_0)\Gamma(N_++\alpha_1)}{\Gamma(N+\alpha_0+\alpha_1)}\\
\frac{\lambda^{m^*}}{m^*!} 
&\geq &
\frac{\Gamma(N_-+\alpha_0)\Gamma(N_++\alpha_1)}{\Gamma(N+\alpha_0+\alpha_1)} 
\prod\limits_{j=0}^{m^*} \left({\frac{\sum\limits_{k \in R_{j-1}(c_{<j},\mathcal{A})}  (\eta^k/k!)}{(\eta^{c_j}/c_j!)}} |Q_{c_j}|\right)
\prod\limits_{j=1}^{m^*} \frac{ \Gamma( N_{j,0}+N_{j,1}+\alpha_0+\alpha_1)}{ \Gamma (N_{j,0}+\alpha_0)\Gamma (N_{j,1}+\alpha_1)}\\
\frac{\lambda^{m^*}}{m^*!} 
&\geq &
\frac{\Gamma(N_-+\alpha_0)\Gamma(N_++\alpha_1)}{\Gamma(N+\alpha_0+\alpha_1)} 
\prod\limits_{j=1}^{m^*} (1 \times |Q_{c_j}|)
\prod\limits_{j=1}^{m^*} 1 \\
\frac{\lambda^{m^*}}{m^*!} 
&\geq &
\frac{\Gamma(N_-+\alpha_0)\Gamma(N_++\alpha_1)}{\Gamma(N+\alpha_0+\alpha_1)} 
\prod\limits_{j=1}^{m^*} |Q_{c_j}|. 
\end{eqnarray*}
By construction we have $\prod\limits_{j=1}^{m^*} |Q_{c_j}| \geq \prod\limits_{j=1}^{m^*}b_j$, thus \[
\frac{\lambda^{m^*}}{m^*!} 
\geq 
\frac{\Gamma(N_-+\alpha_0)\Gamma(N_++\alpha_1)}{\Gamma(N+\alpha_0+\alpha_1)} 
\prod\limits_{j=1}^{m^*} b_j.
\]
We need only the first $m$ terms of the $b_j$'s, the rest are not needed. Note that the left hand side decreases rapidly after $m$ exceeds $\lambda$. In addition to this inequality, there is an additional (trivial) upper limit for $m$, namely the value $2^P-1$, which corresponds to a rule list that includes all of the possible rules. So the length of the optimal rule list should satisfy the following upper bound:\[
m^* \leq m_{\max}=\min \left\{ 2^P-1, \max \left\{ m' \in \mathbb{Z}_+ : \frac{\lambda^{m'}}{m'!} 
\geq 
\frac{\Gamma(N_-+\alpha_0)\Gamma(N_++\alpha_1)}{\Gamma(N+\alpha_0+\alpha_1)} 
\prod\limits_{j=1}^{m'} b_j\right\}          \right\}.
\]
\qed

\section{Proof of Theorem \ref{Theorem_PrefixBound}}
Recall the definition of $N_{j,0}$ as the number of points captured by rule $j$ with label 0, and $N_{j,1}$ as the number of points captured by rule $j$ with label 1,
\[
N_{j,0} = |\{i:\Captr(i)=j \textrm{ and } y_i=0\}|,\;\;N_{j,1}=|\{i:\Captr(i)=j \textrm{ and } y_i=1\}|.
\]
\begin{definition}
For rule $j$, if either  $N_{j,0}$ or $N_{j,1}$ equals zero, rule $j$ is called a \textbf{perfect rule} with respect to $d$. 
\end{definition}
A perfect rule correctly classifies all observations it captures.

\begin{lemma}\label{LemmaHypothetical}
For rule list
\begin{align*}
d = {d_p, a_{p+1}, a_{p+2}, ...,a_j, ..., a_m, a_0}
\end{align*}
where $a_j$ is not a perfect rule, consider a hypothetical rule list
\begin{align*}
d^{\fixed} = {d_p, a_{p+1}, a_{p+2}, ...,a^{j^{+}}, a^{j^{-}},..., a_m, a_0}
\end{align*}
where $a^{j^{+}}$ and $a^{j^{-}}$ are perfect rules with label 1's and 0's, respectively, that capture the same observations as rule $j$, so that $N_{j^{+},1}=N_{j,1}$, $N_{j^{+},0}=0$, and $N_{j^{-},1}=0$, $N_{j^{-},0}=N_{j,0}$. Then, for parameters $\alpha_0=1$ and $\alpha_1=1$,
\begin{align*}
\likelihood(d, \{(x_i,y_i)\}_{i=1}^n) < \likelihood(d^{\fixed}, \{(x_i,y_i)\}_{i=1}^n). 
\end{align*} 
\end{lemma}
(Note that $a^{j^{+}}$ and $a^{j^{-}}$ may not exist in practice, but we create them in theory for the purposes of this proof.)

Intuitively, Lemma \ref{LemmaHypothetical} states that if rule $j$ is not a perfect rule with respect to $d$, meaning $N_{j,0}\geq 1$ and $N_{j,1}\geq 1$, then replacing rule $j$ with two perfect rules that capture the same data points would improve the likelihood.\\

\begin{proof}
We compare the likelihood ratio of the rule lists before and after splitting rule $j$ into two perfect rules.
Splitting the rule will not affect the data points captured by other rules. The likelihood of a rule list is a product of likelihoods for individual rules. Thus,

\begin{eqnarray*}
\lefteqn{\frac{\likelihood(d^{\fixed}, \{(x_i,y_i)\}_{i=1}^n)}{\likelihood(d, \{(x_i,y_i)\}_{i=1}^n)}}\\
&=& \frac{\frac{\Gamma(N_0+1)\Gamma(1)}{\Gamma(N_0+2)} \frac{\Gamma(1)\Gamma(N_1+1)}{\Gamma(N_1+2)}}{\frac{\Gamma(N_0+1)\Gamma(N_1+1)}{\Gamma(N_0+N_1+2)}}
= \frac{(N_0+N_1+1)!}{(N_0+1)!(N_1+1)!} \;\;\;\textrm{(eliminated common factors)}\\
&=& \frac{\binom{N_0+N_1+1}{N_0+1}}{N_1+1}\;\;\; \left(\textrm{using identity }\binom{n}{k} = \frac{n!}{k!(n-k)!}\right)\\
&=& \frac{\binom{N_0+(N_1-1)+1}{N_0+1} + \binom{N_0+N_1}{N_0}}{N_1+1}\;\;\;\left(\textrm{using identity } \binom{n}{k} = \binom{n-1}{k} + \binom{n-1}{k-1}\right) \\
&=& \frac{\binom{N_0+(N_1-1)+1}{N_0+1} + \binom{N_0+N_1}{N_1}}{N_1+1}\;\; \left(\textrm{using identity } \binom{n}{k} = \binom{n}{n-k}\right)
\\
&\geq& \frac{\binom{N_0+1}{N_0+1} + \binom{1+N_1}{N_1}}{N_1+1} \;\;(\textrm{because } N_0, N_1 \geq 1)\\
&=&  \frac{N_1+2}{N_1+1} \\
&>&\ 1.
\end{eqnarray*}
\qed
\end{proof}

Let us discuss the next result, Lemma \ref{Lemma43}. If $j$ and $k$, where $j\leq{k}$ are both perfect rules in $d$ and capture data points with only label $l$'s (where $l$ is either 0 or 1), then replacing \(a_j, a_k\) with a single perfect rule \(a_{kj}\) that captures the same data points will improve the likelihood probability. Formally,
\begin{lemma}\label{Lemma43}
For rule list
\begin{align}
d = {d_p, a_{p+1}, a_{p+2}, ...,a_k, ...,a_j,... a_m, a_0}
\end{align}
where $k$ and $j$ are both perfect rules and have the same label $l$, consider a hypothetical rule list 
\begin{align}
d^{\consolidated} = {d_p, a_{p+1}, a_{p+2}, ...,a_{kj},... a_m, a_0}
\end{align}
where $kj$ is a perfect rule that captures all the data points captured by $k$ and $j$. Then
\begin{align}
\likelihood(d, \{(x_i,y_i)\}_{i=1}^n) 
<\likelihood(d^{\consolidated}, \{(x_i,y_i)\}_{i=1}^n). 
\end{align}
\end{lemma}
\begin{proof}:
\begin{eqnarray*}
\lefteqn{\frac{\likelihood(d^{\consolidated}, \{(x_i,y_i)\}_{i=1}^n)}{\likelihood(d, \{(x_i,y_i)\}_{i=1}^n)}}\\
&=&  \frac{\frac{\Gamma(N_{j,l}+N_{k,l}+\alpha_l)\Gamma(\alpha_l)}{\Gamma(N_{j,l}+N_{k,l}+2\alpha_l)}}{\frac{\Gamma(N_{j,l}+\alpha_l)\Gamma(\alpha_l)}{\Gamma(N_{j,l}+2\alpha_l)} \frac{\Gamma(N_{k,l}+\alpha_l)\Gamma(\alpha_l)}{\Gamma(N_{k,l}+2\alpha_l)}} \;\;\;\textrm{ (by definition)} \\
&=& 
\frac{1}{\Gamma(\alpha_l)}
\frac{(N_{j,l}+\alpha_l)(N_{k,l}+\alpha_l)}{(N_{j,l}+N_{k,l}+\alpha_l)}
\frac{(N_{j,l}+\alpha_l+1)(N_{k,l}+\alpha_l+1)}{(N_{j,l}+N_{k,l}+\alpha_l+1)}\cdots
\frac{(N_{j,l}+2\alpha_l-1)(N_{k,l}+2\alpha_l-1)}{(N_{j,l}+N_{k,l}+2\alpha_l-1)}
\\ 
&=&
\frac{1}{\Gamma(\alpha_l)}
\left[\frac{N_{j,l}N_{k,l}+\alpha_lN_{j,l}+\alpha_lN_{k,l}+\alpha_l^2}{N_{j,l}+N_{k,l}+\alpha_l}
\right]\cdots \left[\frac{N_{j,l}N_{k,l}+(2\alpha_l-1) N_{j,l}+(2\alpha_l-1)N_{k,l}+(2\alpha_l-1)^2}{N_{j,l}+N_{k,l}+(2\alpha_l-1)}
\right]\\
&=&\frac{1}{\Gamma(\alpha_l)}\left[\alpha_l + \frac{N_{j,l}N_{k,l}}{N_{j,l}+N_{k,l}+\alpha_l}\right]\cdots\left[2\alpha_l-1 + \frac{N_{j,l}N_{k,l}}{N_{j,l}+N_{k,l}+(2\alpha_l-1)}\right]\\ 
&>& \frac{1}{\Gamma(\alpha_l)}\left[\alpha_l \right]\left[\alpha_l+1 \right]\left[\alpha_l+2 \right]\cdots \left[2\alpha_l-1 \right]
\\
&\geq& 1. 
\end{eqnarray*}
\qed
\end{proof}

\begin{proof}(\textit{Of Theorem \ref{Theorem_PrefixBound}})
Combining Lemma \ref{LemmaHypothetical} and Lemma \ref{Lemma43}, we can get an upper bound for the posterior of rule list $d$ in terms of the first few rules in the list. 
Lemma \ref{LemmaHypothetical} tells us to separate each rule hypothetically into two perfect rules. Lemma \ref{Lemma43} tells us to combine all perfect rules from the same class into a single rule.
After doing this, there are only two rules left, a perfect rule for class label 0 and a perfect rule for class label 1. We conclude that the likelihood of the rule list \( d = \{d_p, a_{p+1}, a_{p+2},..., a_{m}, a_{0}\} \) is at most the likelihood of the rule list 
\[d^{\hyp}=\{d_p, a_{p_0}, a_{p_1}, a_0\},\] 
where $p_0$ is an imaginary perfect rule in $d^{\hyp}$ capturing all remaining data points with label 0's and $p_1$ is an imaginary perfect rule in $d^{\hyp}$ capturing all remaining data points with label 1's. That is:
\[\likelihood(d,\{(x_i,y_i)\}_{i=1}^n))\leq\likelihood(d^{\hyp},\{(x_i,y_i)\}_{i=1}^n)).\]
We compress notation slightly to remove explicit dependence on the data, so we write $\likelihood(d)=\likelihood(d,\{(x_i,y_i)\}_{i=1}^n)$. Also note that the likelihood of the list can be decoupled into terms for each rule, 
\[\likelihood(d)=\prod_{j=1}^m
\frac{\Gamma(N_{j,0}+\alpha_0)\Gamma(N_{j,1}+\alpha_1)}{\Gamma(N_{j,0}+N_{j,1}+\alpha_0+\alpha_1)}=\prod_{j=1}^m \likelihood(\textrm{rule } j), 
\]
which means that the likelihood for rule list $d^{\hyp}$ can be split into likelihood for the first $p$ rules and likelihood for the other rules.
\begin{eqnarray*}
\lefteqn{\likelihood(d^{\hyp},\{(x_i,y_i)\}_{i=1}^n)=}\\ &&\likelihood(d_p,\textrm{data captured by rules in } d_p)\times 
\likelihood(a_{p_0},\textrm{data captured by }a_{p_0})\times\\
&&
\likelihood(a_{p_1},\textrm{data captured by }a_{p_1})
\times \likelihood(a_{0},\textrm{no data}).
\end{eqnarray*}

Next we show 
$\posterior(d)$ $\leq$
$\posterior(d^{\hyp})$ $\leq \Upsilon(d_p, \{(x_i,y_i)\}_{i=1}^n)$. 
We compute:
\begin{eqnarray}\nonumber
\posterior(d) 
&=& \prior(d)\times\likelihood(d) \\\nonumber
&\leq& \prior(d)\times\likelihood(d^{\hyp}) \\\nonumber
&=&
\prior(\textrm{number of rules in }d)\times \prior(\textrm{size of rules in }d) \times\likelihood(d^{\hyp})
\\\nonumber
&=&
\prior(m|\lambda)\times\prior(\textrm{size of rules in } d_p) \times \prior(\textrm{size of rules in } d\backslash d_p)\times\\\label{bigineq}
&&  \likelihood(d_p)\times\likelihood(a_{p_0})\times\likelihood(a_{p_1})\times \likelihood(a_0).
\end{eqnarray}
Let us handle each term of the expression above, starting with the term for the number of rules. The largest value of the prior occurs at the maximum of the Poisson distribution centered at $\lambda$. That would happen if there were $\lambda$ total rules. This could happen if $p\leq \lambda$. If $p> \lambda$, then we cannot have a rule list of size $\lambda$ since we already have too many rules. In that case, we should not add more rules, and the maximum prior occurs when the size of the rule list is $p$. That is,
\[
\prior(m|\lambda)\leq \frac{\lambda^{\max{(p, \lambda)}}/{(\max{(p, \lambda)})}!}{\sum_{j=0}^{|\mathcal{A}|} (\lambda^j/j!)}.
\]
The second term in \eqref{bigineq} is an equality,
\[
\prior(\textrm{size of rules in } d_p)=\left(\prod_{j=1}^p p(c_j|c_{<j},\mathcal{A},\eta) \frac{1}{|Q_{c_j}|}\right).
\]
The third term in \eqref{bigineq} is trivially bounded by 1. The fourth term $\likelihood(d_p)$ can be calculated from the data as usual, simplifying with $\alpha_0=\alpha_1=1$:
\[
\likelihood(d_p)=\prod_{j=1}^p \frac{ \Gamma (N_{j,0}+\alpha_0)\Gamma (N_{j,1}+\alpha_1)}{ \Gamma(  N_{j,0} + N_{j,1} + \alpha_0+\alpha_1)} =
\prod_{j=1}^p \frac{ \Gamma (N_{j,0}+1)\Gamma (N_{j,1}+1)}{ \Gamma(  N_{j,0} + N_{j,1} + 2)}
\]
For the terms for hypothetical rules $a_{p_0}$ and $a_{p,1}$ we compute them as if those rules were real rules: 
\begin{eqnarray*}
\likelihood(a_{p,0})&=&\frac{ \Gamma (1+N_0-\sum_{j=1}^{p} N_{j,0}) }{ \Gamma(2+N_0-\sum_{j=1}^{p} N_{j,0})}\\
\likelihood(a_{p,1})&=&\frac{ \Gamma (1+N_1-\sum_{j=1}^{p} N_{j,1}) }{ \Gamma(2+N_1-\sum_{j=1}^{p} N_{j,1})}. 
\end{eqnarray*}
The last term of \eqref{bigineq} will be trivially upper bounded by 1. Multiplying all of these terms together to form an upper bound, we have precisely the definition of $\Upsilon(d_p, \{(x_i,y_i)\}_{i=1}^n)$. Thus, 
\[
\posterior(d)\leq \Upsilon(d_p, \{(x_i,y_i)\}_{i=1}^n).
\]
By the assumption of Theorem \ref{Theorem_PrefixBound}, we know that for our rule list $d$, 
\begin{eqnarray*}
\posterior(d)&\leq& \Upsilon(d_p)<v_t^*=\max_{t'\leq t} \posterior(d^{t'},\{(x_i,y_i)\}_{i=1}^n)\leq \max_{d'} \posterior(d'),
\end{eqnarray*}
and more simply stated,
\begin{eqnarray*}
\posterior(d)&<&\max_{d'}\posterior(d').
\end{eqnarray*}
Thus, there is no possible way that our current rule list $d$ could be within $\textrm{argmax}_{d'}\posterior(d')$.\qed
\end{proof}

\bibliographystyle{theapa}

\end{document}